\documentclass[10pt,twocolumn,letterpaper]{article}

\usepackage{wacv}
\usepackage{times}
\usepackage{epsfig}
\usepackage{graphicx}
\usepackage{amsmath}
\usepackage{amssymb}
\usepackage{booktabs}

\usepackage{caption}
\usepackage{subcaption}
\graphicspath{{figs/}{./}} 
\usepackage{amsmath}

\DeclareMathOperator*{\argmin}{arg\,min}
\usepackage[accsupp]{axessibility}  
\usepackage{authblk}

\def\wacvPaperID{1523} 

\wacvalgorithmstrack   

\wacvfinalcopy 

\ifwacvfinal
\usepackage[breaklinks=true,bookmarks=false]{hyperref}
\else
\usepackage[pagebackref=true,breaklinks=true,colorlinks,bookmarks=false]{hyperref}
\fi

\begin{document}

\title{Recovering Fine Details for Neural Implicit Surface Reconstruction}


\makeatletter
\renewcommand\AB@affilsepx{\;\;\;\;\; \protect\Affilfont}
\makeatother

\author[1]{Decai Chen}
\author[1,2]{Peng Zhang}
\author[1]{Ingo Feldmann}
\author[1]{Oliver Schreer}
\author[1,3]{Peter Eisert}
\affil[1]{Fraunhofer HHI} 
\affil[2]{TU Berlin}
\affil[3]{HU Berlin}
\setcounter{Maxaffil}{0}

\maketitle
\thispagestyle{empty}

\begin{abstract}
Recent works on implicit neural representations have made significant strides. Learning implicit neural surfaces using volume rendering has gained popularity in multi-view reconstruction without 3D supervision. However, accurately recovering fine details is still challenging, due to the underlying ambiguity of geometry and appearance representation. In this paper, we present D-NeuS, a volume rendering-base neural implicit surface reconstruction method capable to recover fine geometry details, which extends NeuS by two additional loss functions targeting enhanced reconstruction quality. First, we encourage the rendered surface points from alpha compositing to have zero signed distance values, alleviating the geometry bias arising from transforming SDF to density for volume rendering. Second, we impose multi-view feature consistency on the surface points, derived by interpolating SDF zero-crossings from sampled points along rays. Extensive quantitative and qualitative results demonstrate that our method reconstructs high-accuracy surfaces with details, and outperforms the state of the art.
\footnote {Code: \url{https://github.com/fraunhoferhhi/D-NeuS}.}
\end{abstract}

\section{Introduction}
\begin{figure*}[htbp] 
\centering
\begin{minipage}[t]{0.19\linewidth} 
\centering
\includegraphics[width=0.95\linewidth]{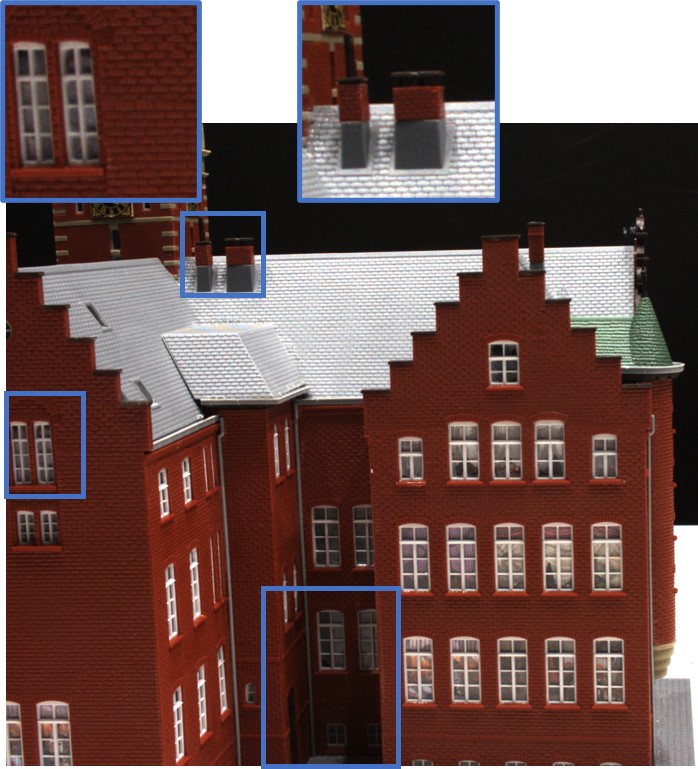} 
\caption*{\label{fig1}{\small Reference Image}} 
\end{minipage}%
\begin{minipage}[t]{0.19\linewidth}
\centering
\includegraphics[width=0.95\linewidth]{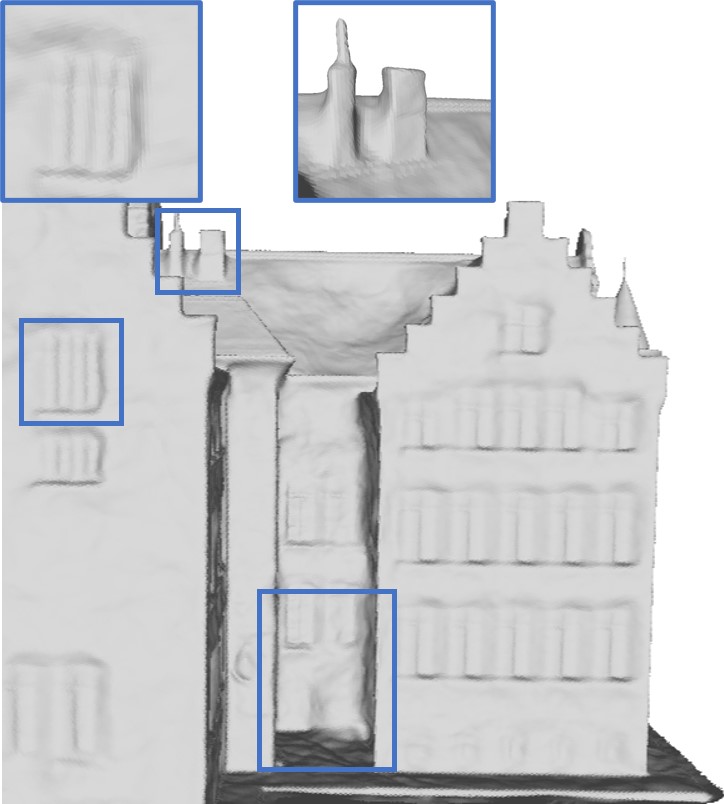}
\caption*{\label{fig2}{\small NeuS~\cite{neus}}}
\end{minipage}%
\begin{minipage}[t]{0.19\linewidth}
\centering
\includegraphics[width=0.95\linewidth]{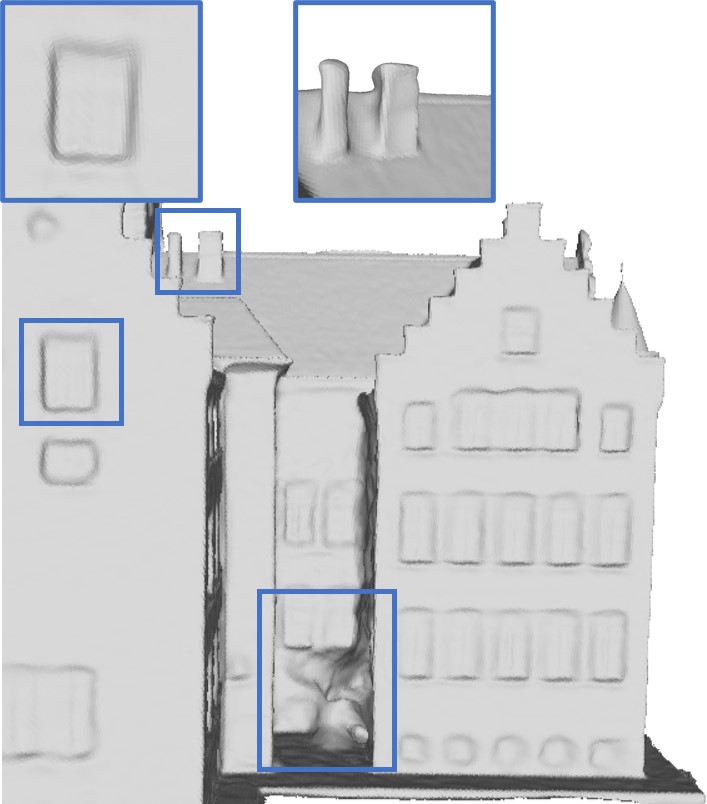}
\caption*{\label{fig3}{\small NeuralWarp~\cite{NeuralWarp}}}
\end{minipage}
\begin{minipage}[t]{0.19\linewidth}
\centering
\includegraphics[width=0.95\linewidth]{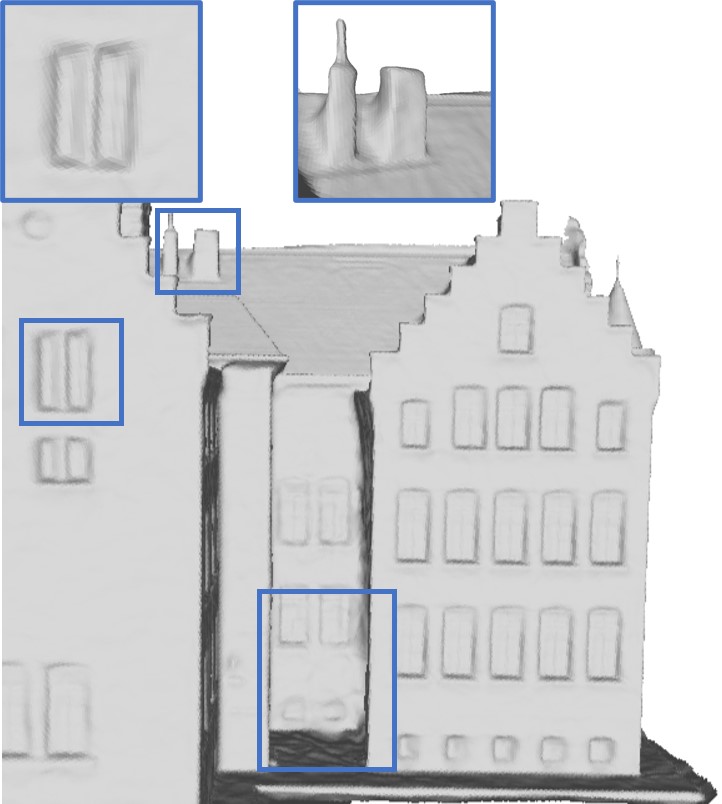}
\caption*{\label{fig4}{\small D-NeuS (Ours)}}
\end{minipage}
\begin{minipage}[t]{0.19\linewidth}
\centering
\includegraphics[width=0.95\linewidth]{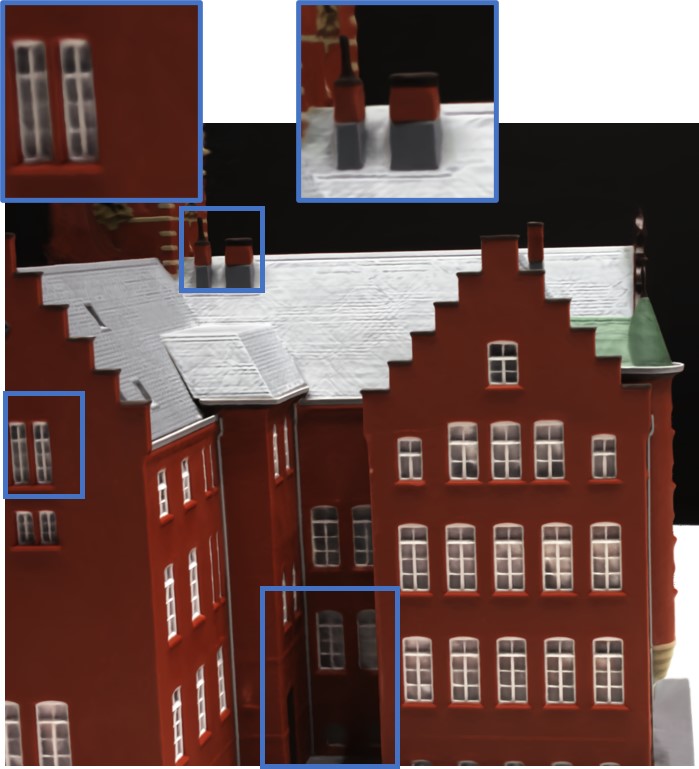}
\caption*{\label{fig5}{\small Our Rendered Image}}
\end{minipage}
\caption{A surface reconstruction example from DTU dataset~\cite{dtu}. Compared to the state-of-the-art methods, D-NeuS recovers higher-fidelity surface details. Besides, we achieve photorealistic view synthesis.}
\label{teaser}
\end{figure*}

3D reconstruction from calibrated multi-view images is a long-standing challenge in computer
vision and has been explored for decades. Classical approaches such as traditional~\cite{galliani2015massively,colmap,Xu2019} and learning-based multi-view stereo (MVS)~\cite{yao2018mvsnet,yao2019recurrent,zhang2020visibilityaware,wang2020patchmatchnet,cao2022mvsformer} produce depth maps via matching photometric or feature correspondence of pixels or patches across a set of images. However, the classical MVS pipeline involves several steps including depth map prediction, fusion into global space and surface extraction, where errors and artifacts inevitably accumulate. Inspired by the seminal work NeRF~\cite{nerf}, neural implicit surface reconstruction approaches~\cite{idr,unisurf,volsdf,neus} have recently emerged as a powerful tool for the 3D reconstruction of geometry and appearance, leveraging coordinate-based Multi-Layer Perceptron (MLP) neural networks to represent the underlying surface. These approaches apply differentiable rendering to optimize jointly the shape and the appearance field by minimizing the gap between rendered images and the input ground truth. While rendering plausible novel views, these methods still struggle to recover high-fidelity geometry details. In this paper, we propose a Details recovering Neural implicit Surface reconstruction method named D-NeuS, with two constraints to guide the SDF field-based volume rendering and thus improve the reconstruction quality. As shown in Fig.~\ref{teaser}, our method is able to reconstruct more accurate geometry details than the state of the art~\cite{neus,NeuralWarp}.

To get rid of geometric errors of the standard volume rendering approaches, NeuS~\cite{neus} applies a weight function which is occlusion-aware and unbiased in the first-order approximation of SDF. However, we argue that the weight function under non-linearly distributed SDF field causes bias between the geometric surface point (\ie SDF zero crossing) and rendered surface point from alpha compositing. To this end, we propose a novel scheme to mitigate this bias. Specifically, we generate additional distance maps during the volume rendering, back-project the distance into 3D points, and penalize their absolute SDF values predicted by the geometry MLP network. By doing this, we encourage the consistency between volume rendering and underlying surface.

Although current neural implicit surface networks are able to render plausible images on novel views, encoding high-frequency textures in MLP is still challenging~\cite{NeuralWarp}. To alleviate this issue, NeuralWarp~\cite{NeuralWarp} introduces patch-based photometric consistency, which is computed on all sampled points along rays before merging by alpha compositing. Inspired by MVSDF~\cite{mvsdf}, we instead take the advantage of robust representation performance of convolutional neural networks (CNNs), by employing feature-based consistency on only the surface point along a ray. Instead of finding surface points via ray tracing as in~\cite{mvsdf}, which requires recursively querying the geometry network and thus is computationally expensive, we simply look for the first zero crossing point from the SDF values of the sampled points via locally differentiable linear interpolation. This involves no extra computation because the SDF values of the sampled points are computed already for volume rendering. 

To summarize, the main contributions of our work are as follows:
\begin{itemize}
 \item We provide theoretical analysis of the geometry bias resulting from the unregularized SDF field in volume rendering-based neural implicit surface network, and propose a novel constraint to regularize this bias.
 \item We apply multi-view feature consistency on surface points from linearly interpolated SDF zero-crossing, for fine local geometric details.
 \item We evaluate qualitatively and quantitatively the proposed method on DTU~\cite{dtu}, BlendedMVS~\cite{blendedmvs} datasets, and show that it outperforms the state of the art, with high-accuracy surface reconstruction, especially on complex scenes.
\end{itemize}

\section{Related Works}
\noindent
\textbf{Multi-View Stereo.} MVS is a classical method for recovering a dense scene representation from overlapping images. Traditional MVS approaches typically leverage pair-wise matching cost of RGB image patches by Normalized Cross-Correlation (NCC), Sum of Squared Distances (SSD) or Sum of Absolute Differences (SAD). In recent years, PatchMatch-based~\cite{bleyer2011patchmatch} MVS~\cite{galliani2015massively,colmap,Xu2019} are dominating traditional methods due to highly parallelism and robust performance. Recently, deep learning-based MVS shows superior performance. MVSNet~\cite{yao2018mvsnet} builds cost volumes by warping feature maps from across neighboring views and applies 3D CNNs to regularize the cost volumes. To mitigate the memory consumption of 3D CNNs, R-MVSNet~\cite{yao2019recurrent} regularizes 2D cost maps sequentially using a gated recurrent network, while other methods~\cite{cheng2020deep,yang2020cost,gu2020cascade,zhang2020visibilityaware} integrate coarse-to-fine multi-stage strategies to progressively refine the 3D cost volumes. PatchmatchNet~\cite{wang2020patchmatchnet} proposes an iterative multiscale PatchMatch strategy in a differentiable MVS architecture. More recently, TransMVSNet~\cite{ding2022transmvsnet} introduces Transformer to aggregate long-range context information within and across images. However, matching pixels in low-texture or non-Lambertian areas remains challenging, and errors inevitably accumulate from the following point cloud fusion and surface reconstruction. In this work, we leverage multi-view feature consistency, universally used in learning-based MVS, to constrain the volume rendering for more accurate surface reconstruction.\\

\begin{figure*}
\begin{center}
\includegraphics[trim={0cm 0cm 0cm 0cm},clip,width=\linewidth]{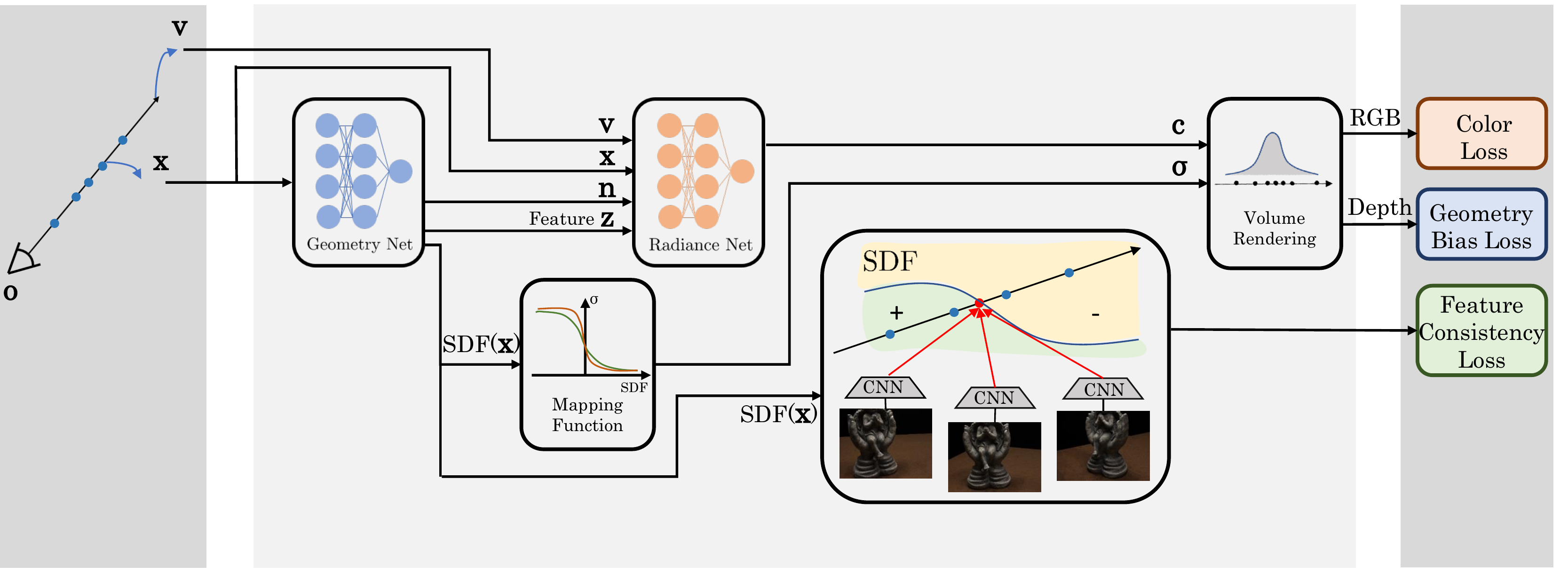}
\end{center}
   \caption{Overview of the proposed method. We build on a neural implicit surface framework~\cite{neus} and introduce two additional constraints: a geometry bias loss which regularizes SDF-based volume rendering (Sec.~\ref{bias}), and multi-view feature consistency loss(Sec.~\ref{feature}), to significantly improve the reconstruction quality.}
\label{fig:pipeline}
\end{figure*}

\noindent
\textbf{Implicit Surface Representation and Reconstruction.} The success of NeRF~\cite{nerf} in representing a scene by 5D radiance field has recently drawn considerable attention from the community of both computer vision and computer graphics. Implicit neural representation leverage physics-based traditional volume rendering in a differentiable way, enabling photorealistic novel view synthesis without 3D supervision. While NeRF-like approaches~\cite{nerf,zhang2020nerf++,barron2022mipnerf360} achieve impressive rendering quality, their underlying geometry is generally noisy and less favorable. The reason for that is two-fold. First, the geometry and appearance fields are entangled in differentiable rendering, if only learned from 2D image reconstruction consistency. Secondly, representing the geometry field by density is difficult to be constrained and regularized. 

To alleviate the above issue, current implicit surface reconstruction methods employ surface indicator functions, mapping continuous spatial coordinates to occupancy~\cite{Occupancy, peng2020convolutional, dvr, unisurf} and SDF~\cite{DeepSDF,idr,neus,volsdf}, where Marching cubes~\cite{lorensen1987marching} is commonly applied to extract the implicit surface at any resolution. IDR~\cite{idr} renders the color of a ray only on the object surface point, and applies differentiable ray tracing to back-propagate the gradients to a local region near the intersection. MVSDF~\cite{mvsdf} extends this framework with supervision from depth maps and feature consistency, while RegSDF~\cite{RegSDF} introduces supervision by point clouds and additional geometric regularization to reconstruct unbounded or complex scenes. However, surface rendering-based methods struggle with reconstructing complex objects with sudden depth changes, and thus they usually require additional supervision, such as object masks, depth maps or point clouds. 

To combine advantages of surface-based and volume-based rendering techniques, UNISURF~\cite{unisurf} proposes a coarse-to-fine strategy for point sampling around the surface represented by an occupancy field. VolSDF~\cite{volsdf} trains an implicit surface model using an efficient sampling algorithm, guided by error bound of opacity approximation. NeuralWarp~\cite{NeuralWarp} extends VolSDF by adding patch warping from source images to the reference image using homography, and improve the surface geometry by photometric constraint. Since patch warping requires reliable surface normals, NeuralWarp only serves as a post-processing method to fine-tune and optimize a pretrained surface model. Similar to VolSDF, NeuS~\cite{neus} designs an occlusion-aware transformation function mapping signed distances to weights for volume rending, with a learnable parameter to control the slope of the logistic density function. However, this mapping function is only unbiased in a regularized SDF field which is linearly distributed, so we propose a novel constraint to compensate for the geometry bias. We build our framework on NeuS~\cite{neus}, but we believe our proposed method could be adapted to any volume rendering-based neural implicit surface reconstruction work.

\section{Method}
Given a set of images with known intrinsic and extrinsic camera parameters, the goal of our method is to reconstruct high-fidelity surface represented by implicit neural networks. Following NeuS~\cite{neus}, we encode surfaces as signed distance fields. The overview of our framework is illustrated in Fig.~\ref{fig:pipeline}. We will explain our method in four parts: 1) First, we show how SDF-based neural implicit surfaces are learned via volume rendering (Section~\ref{review}). 2) Then, we analyze the geometry bias of volume rendering in unregularized SDF field and propose a novel constraint to mitigate this error (Section~\ref{bias}). 3) We demonstrate how to apply feature consistency on surface points from linearly interpolated SDF zero-crossings (Section~\ref{feature}). 4) Finally, we present all losses used for optimization (Section~\ref{loss}).

\subsection{Volume Rendering-based Implicit Surface Reconstruction} \label{review}
In this section, we review the basics of SDF-based neural surface reconstruction using volume rendering~\cite{neus}, so that we can better demonstrate our analysis in Section~\ref{bias}. In contrast to the density-based geometry representation, the surface is represented implicitly by the zero-level set of the SDF field. The surface $S$ of the object is represented by the zero-set of the implicit signed distance field, that is
\begin{equation}
   \mathcal{S} = {\mathbf{x}\in \mathbb{R}^{3}\mid f(\mathbf{x})=0}.
\end{equation}
where $f$ is a function $f:\mathbb{R} ^{3} \rightarrow \mathbb{R}$ mapping a 3D point $\mathbf{x}\in \mathbb{R}^{3}$ to its SDF field. In addition to geometry, we represent view-dependent appearance field by a function $g:\mathbb{R}^{3}\times\mathbb{S}^{2}\times\mathbb{S}^{2}\times\mathbb{R}^{m} \rightarrow \mathbb{R}^{3}$ that encodes the color $\mathbf{c}\in \mathbb{R}^{3}$ associated with a point $\mathbf{x}\in \mathbb{R}^{3}$, its view direction $\mathbf{v}\in \mathbb{S}^{2}$, its normal $\mathbf{n}\in \mathbb{S}^{2}$ calculated from automatic differentiation of the SDF (\ie, $\nabla f(\mathbf{x})$), and a feature vector $\mathbf{z}\in \mathbb{R}^{N_f}$ from the geometry network $f$, as shown in Fig.~\ref{fig:pipeline}. Both functions are encoded by Multi-layer Perception (MLP) neural networks. 

A 3D point emitted from a ray through the corresponding pixel can be denoted as:
\begin{equation}
    \mathbf{x}(t) = \mathbf{o}+t\mathbf{v} \mid t \ge0,
    \label{3dpoint}
\end{equation}
where $\mathbf{o}$ is the camera center, $\mathbf{v}$ is the unit direction vector of the ray, and $t$ is the distance between $\mathbf{x}$ and $\mathbf{o}$. Colors along a ray are accumulated by volume rendering
\begin{equation}
  C(\mathbf{o},\mathbf{v}) = \int_{0}^{+\infty } \omega (t)g(\mathbf{x}(t),\mathbf{v},\mathbf{n},\mathbf{z})dt,
\end{equation}
where $C(\mathbf{o},\mathbf{v})$ is the output color for the pixel, $g(\mathbf{x}(t),\mathbf{v},\mathbf{n},\mathbf{z})$ is the color of a point $\mathbf{x}$ along the view direction $\mathbf{v}$, $\omega(t)$ is the weight for volume rendering at the point:
\begin{equation}
    \omega(t)=exp\left(-\int_{0}^{t}\sigma(u)du\right)\sigma(t),
    \label{eq_weight}
\end{equation}
where $\sigma(t)$ is the density of the point $\mathbf{x}$ used in standard volume rendering. After rendering a set of rays, the rendered colors are compared with the input images for network supervision.

\subsection{Constraint on Geometric Bias} \label{bias}
One key to learn an SDF representation from 2D images is to build an appropriate connection between the volume-rendered colors and SDF values, \ie, to derive an appropriate density or weight function $\omega(t)$ on the ray based on the SDF $f$ of the scene. Assuming the signed distance field is a linear function near a surface point, which is the first-order approximation of SDF, NeuS~\cite{neus} proposed an unbiased opaque density function:
\begin{equation}
  \sigma(t)=\max\left(\frac{-\frac{d\Phi _{s} }{dt}(f(\mathbf{x}(t)))  }{\Phi _{s}(f(\mathbf{x}(t))) },0\right),
  \label{eq_density}
\end{equation}
where $\Phi_{s}(x)=(1+e^{-sx})^{-1}$ is a sigmoid function and $s^{-1}$ is the trainable standard deviation which approaches 0 as training converges.

\begin{figure}[t]
     \centering
     \begin{subfigure}[b]{0.49\columnwidth}
         \centering
         \includegraphics[width=\textwidth]{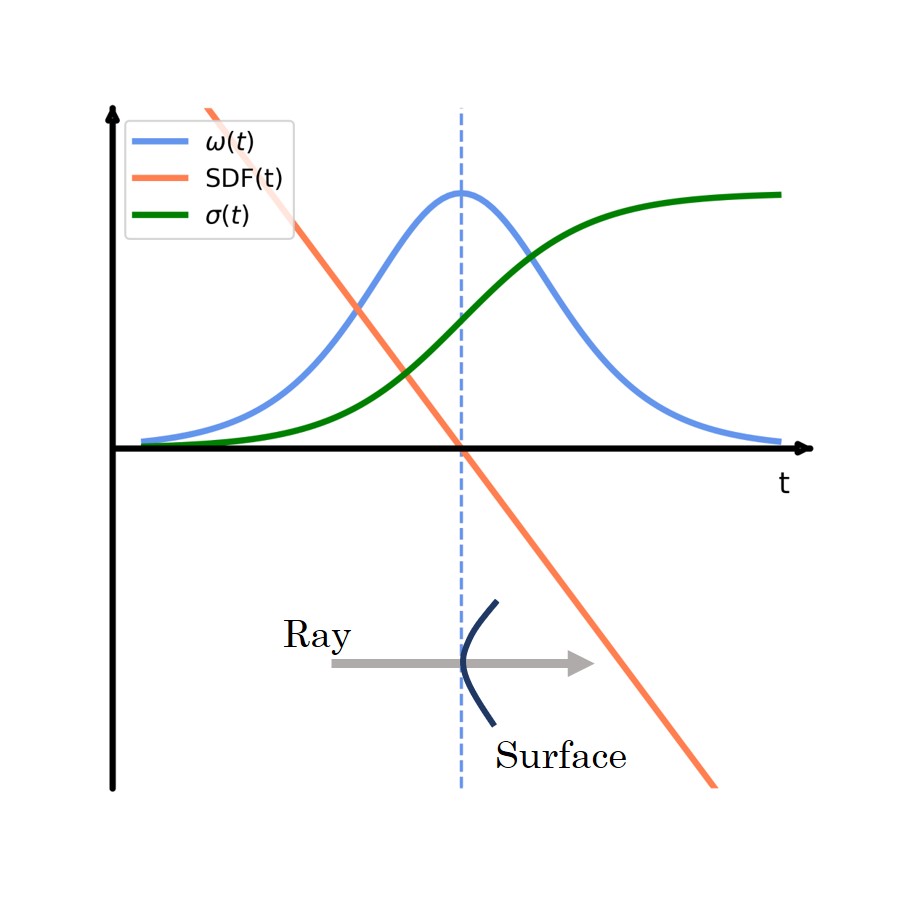}
         \caption{Ideal SDF}
         \label{fig:linearsdf}
     \end{subfigure}
    \hspace*{\fill}
     \begin{subfigure}[b]{0.49\columnwidth}
         \centering
         \includegraphics[width=\textwidth]{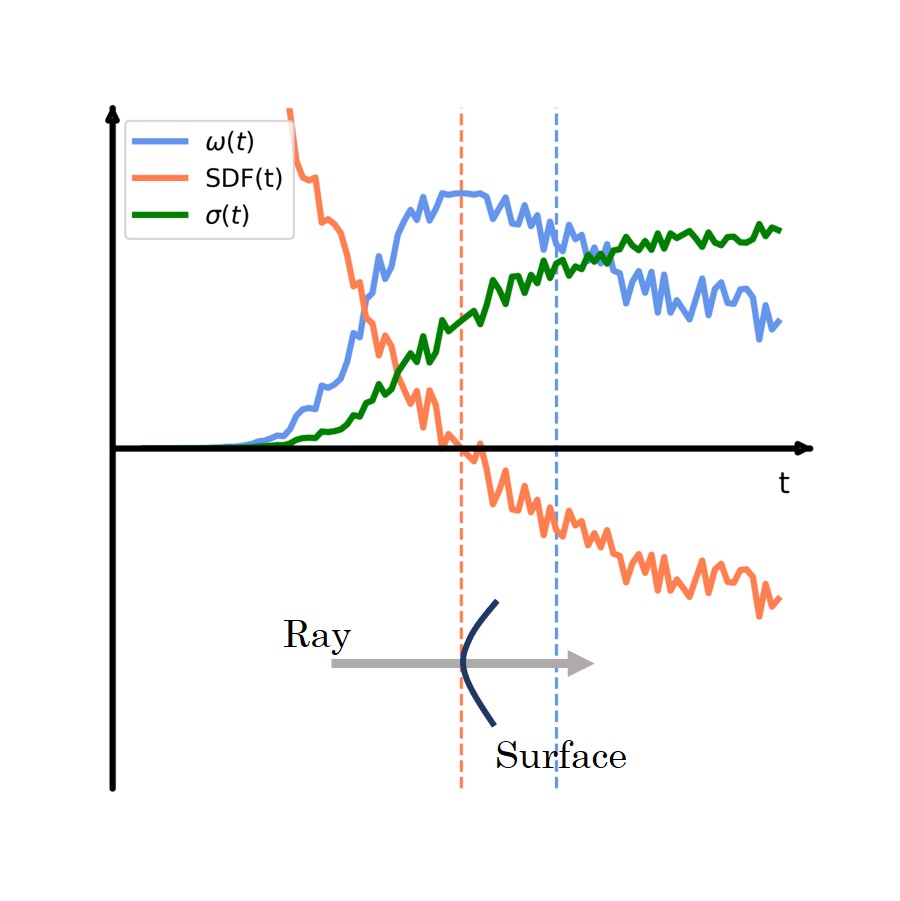}
         \caption{Unregularized SDF}
         \label{fig:irregularsdf}
     \end{subfigure}
        \caption{Illustration of the density and weight functions under different assumptions of SDF distribution. Non-linear SDF values cause bias between geometric surface (orange dashed line) and volume rendered surface point (blue dashed line).}
        \label{fig:sdf_weight}
\end{figure}

Fig.~\ref{fig:sdf_weight} shows how density and weight functions behave under different SDF distributions, in the simple case of a single plane intersection. Assuming the local surface as a plane, the ideal SDF value of a sampling point near the surface is linear along the camera ray, i.e., $f(\mathbf{x}(t)) = -|cos(\theta)|\cdot (t-t^*)$, where $f(\mathbf{x}(t^*) = 0$, and $\theta$ is the locally constant angle between the view direction and the outward surface normal. Based on this assumption of SDF, NeuS~\cite{neus} derives the unbiased weight distribution using Eqn.~\ref{eq_weight}, as demonstrated in Fig.~\ref{fig:linearsdf}. In this case, the point corresponding to the weighted average in volume rendering shares the same position where the SDF value is 0. In other words, the rendered color is consistent with the underlying geometry, so the supervision from input images can optimize the surface geometry precisely.

However, the ideal SDF distribution is not guaranteed by the geometry MLP network, which takes a 3D point $\mathbf{x}$ and outputs its signed distance to the nearest surface. Although the weights of the geometry network are initialized to produce an approximate SDF of a unit sphere~\cite{sdfinit}, the volume rendering-based image supervision itself imposes no explicit regularization on the underlying SDF field. Fig.~\ref{fig:irregularsdf} illustrates an example of unregularized SDF distribution along a camera ray, causing bias between the volume rendering integral and the SDF implicit surface. As a result, the inconsistency between color radiance field and geometry SDF field leads to less desirable surface reconstruction. 

At this point, we propose a novel strategy to regularize the SDF field for volume rendering by constraining the mentioned geometry bias. Recalling a 3D point along a ray in Eqn.~\ref{3dpoint}, we can render the distance $t_{rendered}$ between the camera center and the average point for volume rendering via discretizing the volume integration:
\begin{equation}
 t_{rendered} = \sum_{i}^{n} \frac{\omega _{i} t_{i}}{\sum_{i}^{n}\omega _{i} },
\end{equation}
where n is the number of sampling points along a ray, $\omega _{i}$ represents the discrete counterpart of the weight in Eqn.~\ref{eq_weight}, and $t_{i}$ is the distance from a sampling point to the camera center. Then the volume-rendered surface point $\mathbf{x}_{renderd}$ can be formed by back-projection:
\begin{equation}
    \mathbf{x}_{rendered} = \mathbf{o} + t_{rendered}\mathbf{v}.
\end{equation}
Finally, we build a geometry bias loss:
\begin{equation}
    \mathcal{L}_{bias} = \frac{1}{|\mathbb{S}|} \sum_{\mathbf{x}_{rendered} \in \mathbb{S}} |f(\mathbf{x}_{rendered})|,
    \label{l_bias}
\end{equation}
where $f$ is the geometry network outputting SDF values, $\mathbb{S}$ is the subset of $\mathbf{x}_{rendered}$ where ray-surface intersection has been found (see Sec.~\ref{feature} for implementation details). By penalizing the absolute value of SDF of the rendered surface points, we encourage the geometry consistency between the implicit SDF field and the radiance field for volume rendering. Intuitively, this constraint regularizes the SDF distribution for unbiased volume rendering, and thus leads to more accurate surface reconstruction. 
It is also worth noting, that Eikonal loss~\cite{eikonal} widely used in neural implicit surface reconstruction regularizes the gradient field of SDF by constraining the gradient norm. Both Eikonal loss and our geometry bias loss support each other, enhancing the reconstruction quality.


\subsection{Multi-view Feature Consistency} \label{feature}
Guiding geometry reconstruction with multi-view photometric or feature consistency is popular in MVS~\cite{colmap,Xu2019,yao2018mvsnet,zhang2020visibilityaware} and recent neural surface reconstructions~\cite{mvsdf,NeuralWarp}. Typically, photo-consistency approaches compare the photometric distance across RGB image patches requiring surface normals to compute homography, while feature consistency methods match only a single pixel between the feature maps. Extensive results, \eg benchmarks on Tanks and Temples\cite{tntBenchmark,tnt}, demonstrate that the deep feature representation shows better performance than the photometric counterpart. Therefore, we apply feature consistency to impose multi-view geometric constraint on the reconstructed object surface. 

One key step for applying the multi-view constraint on neural implicit surfaces is to find the surface point in a differentiable way. In surface rendering-based neural reconstruction~\cite{idr,mvsdf}, differentiable ray tracing is commonly used to find the intersection point between a camera ray and object surface. However, for optimizing volume rendering-based surface reconstruction, applying ray tracing to find the surface point causes extra calculation, as it is not required for color rendering. Instead, NeuralWarp~\cite{NeuralWarp} approximates the surface point using alpha composition, \ie, calculating the patch transformation on every sampling point along a ray and merging the results by volume weighted average. However, the volume-rendered surface point can be biased against the real surface, as discussed in Sec.~\ref{bias}. To this end, we take the advantage of the SDF values of the sampling points, which are computed already for volume rendering, to directly extract the zero-crossing point using linear interpolation. 

Recalling Eqn.~\ref{3dpoint}, we denote a sampled 3D point along a ray as $\mathbf{x}(t_{i})$ where $i={1,...,N}$ is the index and $N$ is the number of sampled points from hierarchical sampling~\cite{neus}. We search for the first point $\mathbf{x}(t_{s})$, satisfying that the SDF value of this point is positive while that of the next sampling point is negative. Specifically, we can define $s$ as:
\begin{equation}
s = \argmin_i \{t_{i} \mid f(\mathbf{x}(t_{i})) > 0 \: and \: f(\mathbf{x}(t_{i+1})) < 0 \},
\end{equation}
where $f$ is the geometry network outputting SDF values, and $\mathbf{x}(t_{i})$ is the first sampling point right before the object surface. We only consider the first ray-surface intersection because the others are occluded. If none of the sampling points fulfills such requirement, we skip constraints on both the feature consistency and the geometric bias for this ray. Since the hierarchical sampling strategy puts high importance on sampling near the surface point, the distance between $\mathbf{x}(t_{i})$ and $\mathbf{x}(t_{i+1})$ is supposed to be small. Therefore, we can approximate the surface point where SDF is zero using differentiable linear interpolation:
\begin{equation}
\hat{\mathbf{x}} = \left \{ \mathbf{x}(\hat{t}) \mid \hat{t}=\frac{f(\mathbf{x}(t_{s}))t_{s+1} - f(\mathbf{x}(t_{s+1}))t_{s}  }{f(\mathbf{x}(t_{s})) -f(\mathbf{x}(t_{s+1})) } \right \}. 
\end{equation}
It is worth noting that IDR~\cite{idr} employs a similar strategy for ray marching in a surface rendering pattern, using a recursive secant root-finding algorithm in case the sphere tracing method does not converge. In contrast, we reconstruct the surface using volume rendering and directly approximate the ray-surface crossing with only a single iteration of the secant method thanks to the hierarchical sampling strategy. 

After deriving the surface point $\hat{\mathbf{x}}$, we compare features of this point across multiple views. Similar to MVSDF~\cite{mvsdf}, we extract feature from RGB images with a convolutional neural network (CNN) that is pre-trained for supervised MVS~\cite{zhang2020visibilityaware}. Then we constrain the neural implicit surface reconstruction using a multi-view feature consistency loss: 
\begin{equation}
\mathcal{L}_{feat.} = \frac{1}{N_{c}N_{v}}\sum_{i=1}^{N_{v}}\left |\mathbf{F_{0}}(\mathbf{p}_{0})-\mathbf{F}_{i}(\mathbf{K}_{i}(\mathbf{R}_{i}\hat{\mathbf{x}}+\mathbf{t}_{i}))\right|,
\label{featureloss}
\end{equation}\\
where $N_{v}$ and $N_{c}$ are the numbers of neighboring source views and feature channels respectively, $\mathbf{F}$ is the extracted feature map, $\mathbf{p}_{0}$ is the pixel through which the ray casts, $\{\mathbf{K}_{i},\mathbf{R}_{i},\mathbf{t}_{i}\}$ are the camera parameters of the $i$-th source view.

\begin{table*}[h]
\begin{center}
\resizebox{\textwidth}{!}{
\begin{tabular}{c|ccccccccccccccc|c}
 Scan &24&37&40&55&63&65&69&83&97&105&106&110&114&118&122& means\\
\hline 
 IDR~\cite{idr} & 1.63 & 1.87 & 0.63 & 0.48 & 1.04 & 0.79 & 0.77 & 1.33 & 1.16 & 0.76 & 0.67 & 0.90 & 0.42 & 0.51 & 0.53 & 0.90\\
 MVSDF~\cite{mvsdf} & 0.83 & 1.76 & 0.88 & 0.44 & 1.11 & 0.90 & 0.75 & 1.26 & 1.02 & 1.35 & 0.87 & 0.84 & 0.34 & 0.47 & 0.46 & 0.88\\
 NeuS~\cite{neus} & 0.83 & 0.98 & 0.56 & 0.37 & 1.13 & 0.59 & 0.60 & 1.45 & 0.95 & 0.78 & 0.52 & 1.43 & 0.36 & 0.45 & 0.45 & 0.77\\
 RegSDF~\cite{RegSDF} &  0.60 & 1.41 & 0.64 & 0.43 & 1.34 & 0.62 & 0.60 & 0.90 & 0.92 & 1.02 & 0.60 & 0.60 & 0.30 & 0.41 & 0.39 & 0.72\\
 \hline
 COLMAP~\cite{colmap} & 0.81 & 2.05 & 0.73 & 1.22 & 1.79 & 1.58 & 1.02 & 3.05 & 1.40 & 2.05 & 1.00 & 1.32 & 0.49 & 0.78 & 1.17 &  1.36\\
 VolSDF~\cite{volsdf} & 1.14 & 1.26 & 0.81 & 0.49 & 1.25 & 0.70 & 0.72 & \underline{1.29} & 1.18 & \underline{0.70} & 0.66 & 1.08 & 0.42 & 0.61 & 0.55 & 0.86 \\
 NeuS~\cite{neus} & 1.00 & 1.37 & 0.93 & 0.43 & 1.10 & \underline{0.65} & \underline{0.57} & 1.48 & 1.09 & 0.83 & \underline{0.52} & 1.20 & \underline{0.35} & \underline{0.49} & 0.54  & 0.84\\
NeuralWarp~\cite{NeuralWarp} & \underline{0.49} & \textbf{0.71} & \underline{0.38} & \textbf{0.38} & \textbf{0.79} & 0.81 & 0.82 & \textbf{1.20} & \underline{1.06} & \textbf{0.68} & 0.66 & \underline{0.74} & 0.41 & 0.63 & \underline{0.51} & \underline{0.68}\\
D-NeuS (Ours) & \textbf{0.44} & \underline{0.79} & \textbf{0.35} & \underline{0.39} & \underline{0.88} & \textbf{0.58} & \textbf{0.55} & 1.35 & \textbf{0.91} & 0.76 & \textbf{0.40} & \textbf{0.72} & \textbf{0.31} & \textbf{0.39} & \textbf{0.39} & \textbf{0.61}\\
\end{tabular}}
\end{center}
\caption{Quantitative results of the Chamfer distances on DTU dataset (lower values are better). COLMAP results are achieved by trim=0. The upper part of the table are the neural implicit surfaces reconstruction methods that require additional scene-specific prior inputs such as object masks, depth maps or point clouds). The best score for each scan is marked in \textbf{bold} and the second-best one is \underline{underlined}.}
\label{chamfer}
\end{table*}

\subsection{Training Loss} \label{loss}
The overall loss function to train our neural implicit surface reconstruction network is defined as the weighted sum of the following four terms:
\begin{equation}
\mathcal{L} =  \mathcal{L}_{color} + \alpha \mathcal{L}_{eik.} + \beta\mathcal{L}_{bias}+\gamma\mathcal{L}_{feat.}.
\label{totallosss}
\end{equation}
$\mathcal{L}_{color}$ is the difference between the RGB color taken from ground truth input images $C$ and that from volume rendering $\hat{C}$:
\begin{equation}
\mathcal{L}_{color}=\frac{1}{m}\sum_{i=1}^{m}\left |C_{i} - \hat{C_{i}}\right|, 
\label{colorloss}
\end{equation}\\
where m is the number of pixels trained in a batch.
Following previous works~\cite{idr,mvsdf,volsdf,neus,NeuralWarp}, we add an Eikonal loss~\cite{eikonal} on the sampled points to regularize the gradients of SDF field from the geometry network $f$:
\begin{equation}
\mathcal{L}_{eik.} = \frac{1}{|\mathbb{P}|} \sum_{\mathbf{x}\in\mathbb{P}} (\left\Vert \nabla f(\mathbf{x})  \right\Vert_{2} -1)^2,
\end{equation}
where $\mathbb{P}$ is the set of all sampled points in a batch, and $\left\Vert \cdot \right\Vert_{2}$ is the L2 norm.

\begin{figure*}[ht]
    \captionsetup[subfigure]{labelformat=empty}
    \centering
    \hspace*{\fill}
     \rotatebox{90}{\scriptsize{~~~~~~~~~~~~~~~Scan 40}}
    \hspace*{\fill}
     \begin{subfigure}[b]{0.19\linewidth}
         \centering
         \includegraphics[width=\textwidth]{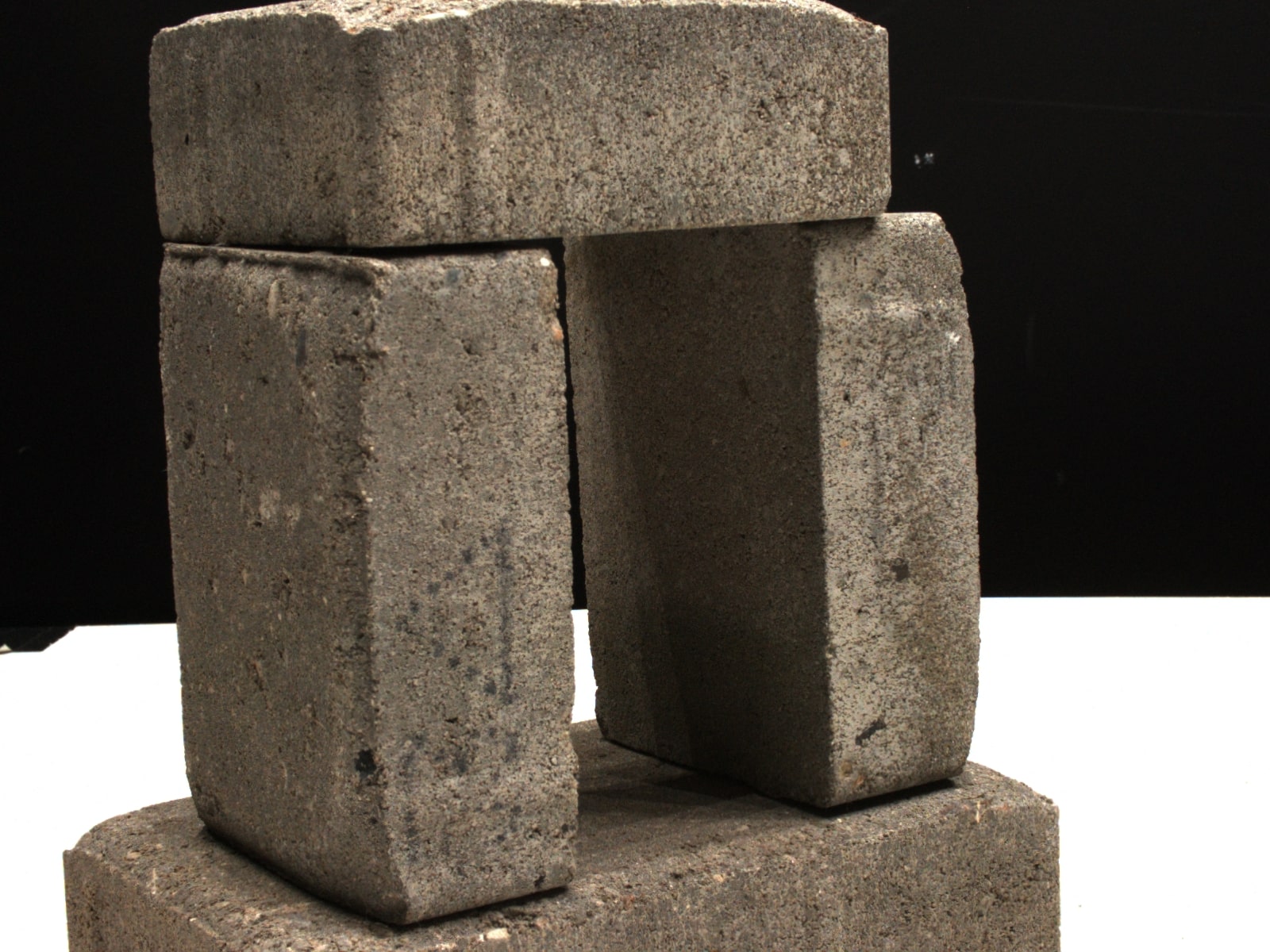}
     \end{subfigure}
     \begin{subfigure}[b]{0.19\linewidth}
         \centering
         \includegraphics[trim={2cm 0 0 0},clip, width=\textwidth]{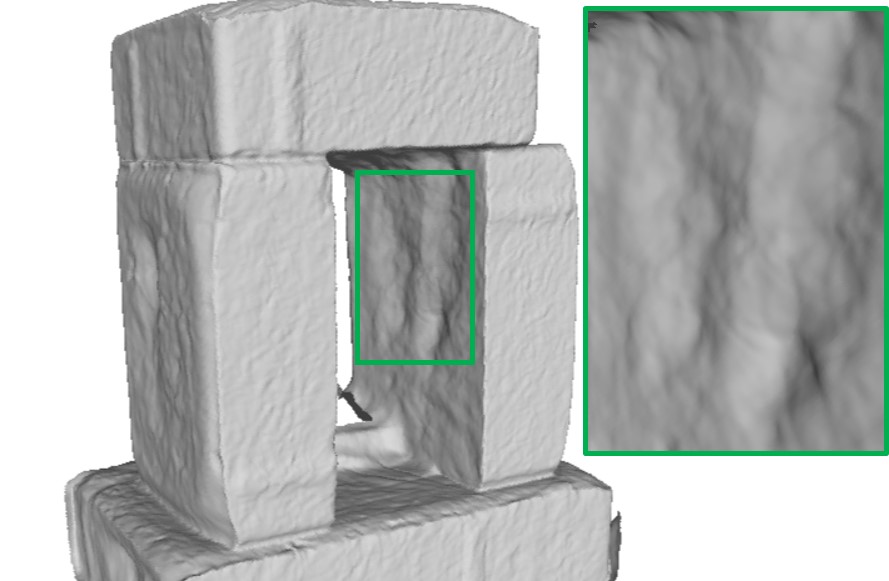}
     \end{subfigure}
     \begin{subfigure}[b]{0.19\linewidth}
         \centering
         \includegraphics[trim={2cm 0 0 0},clip, width=\textwidth]{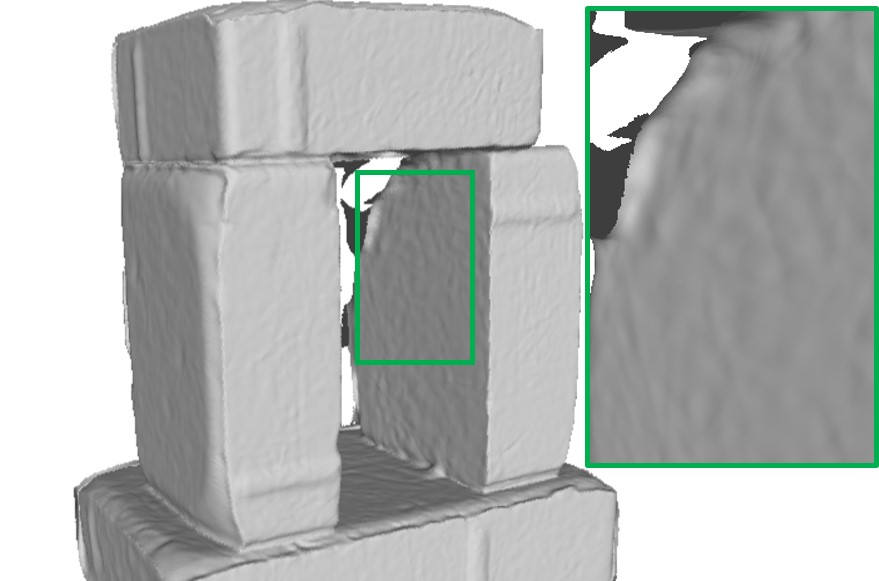}
     \end{subfigure}
     \begin{subfigure}[b]{0.19\linewidth}
         \centering
         \includegraphics[trim={2cm 0 0 0},clip, width=\textwidth]{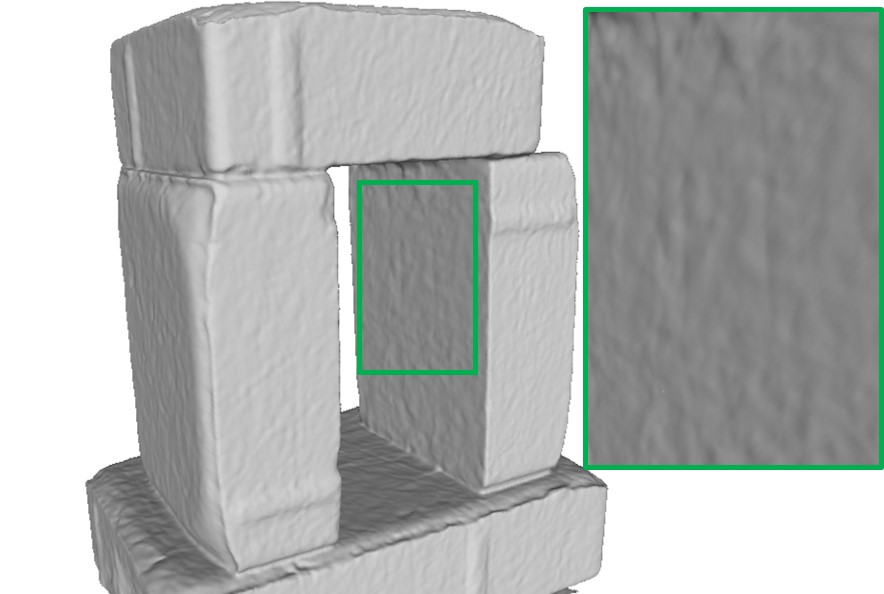}
     \end{subfigure}
     \begin{subfigure}[b]{0.19\linewidth}
         \centering
         \includegraphics[width=\textwidth]{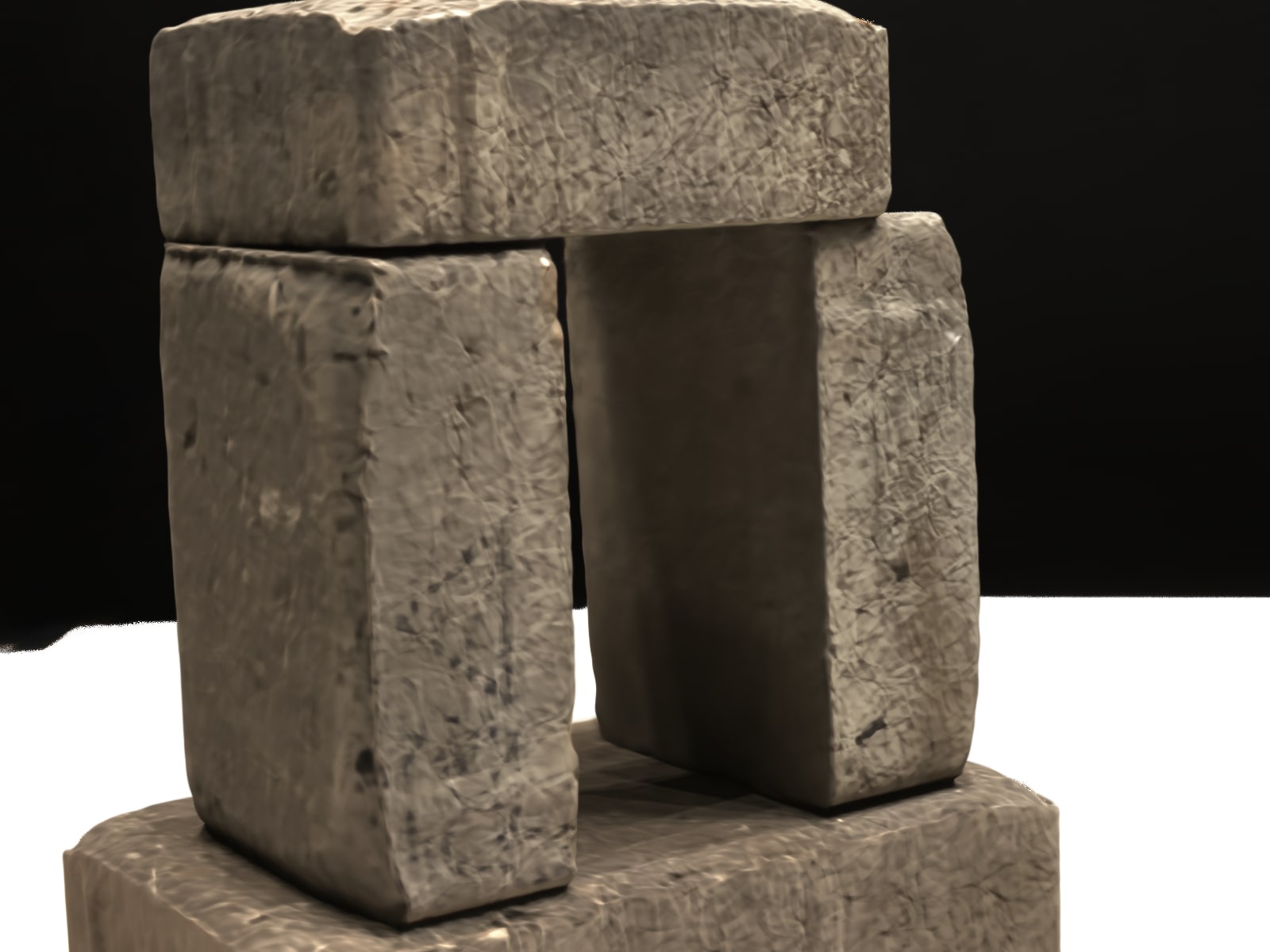}
     \end{subfigure}
    \hspace*{\fill}
    
    \vspace{1.5mm}
    
     \hspace*{\fill}
     \rotatebox{90}{\scriptsize{~~~~~~~~~~~~~Scan 63}}
    \hspace*{\fill}
     \begin{subfigure}[b]{0.19\linewidth}
         \centering
         \includegraphics[width=\textwidth]{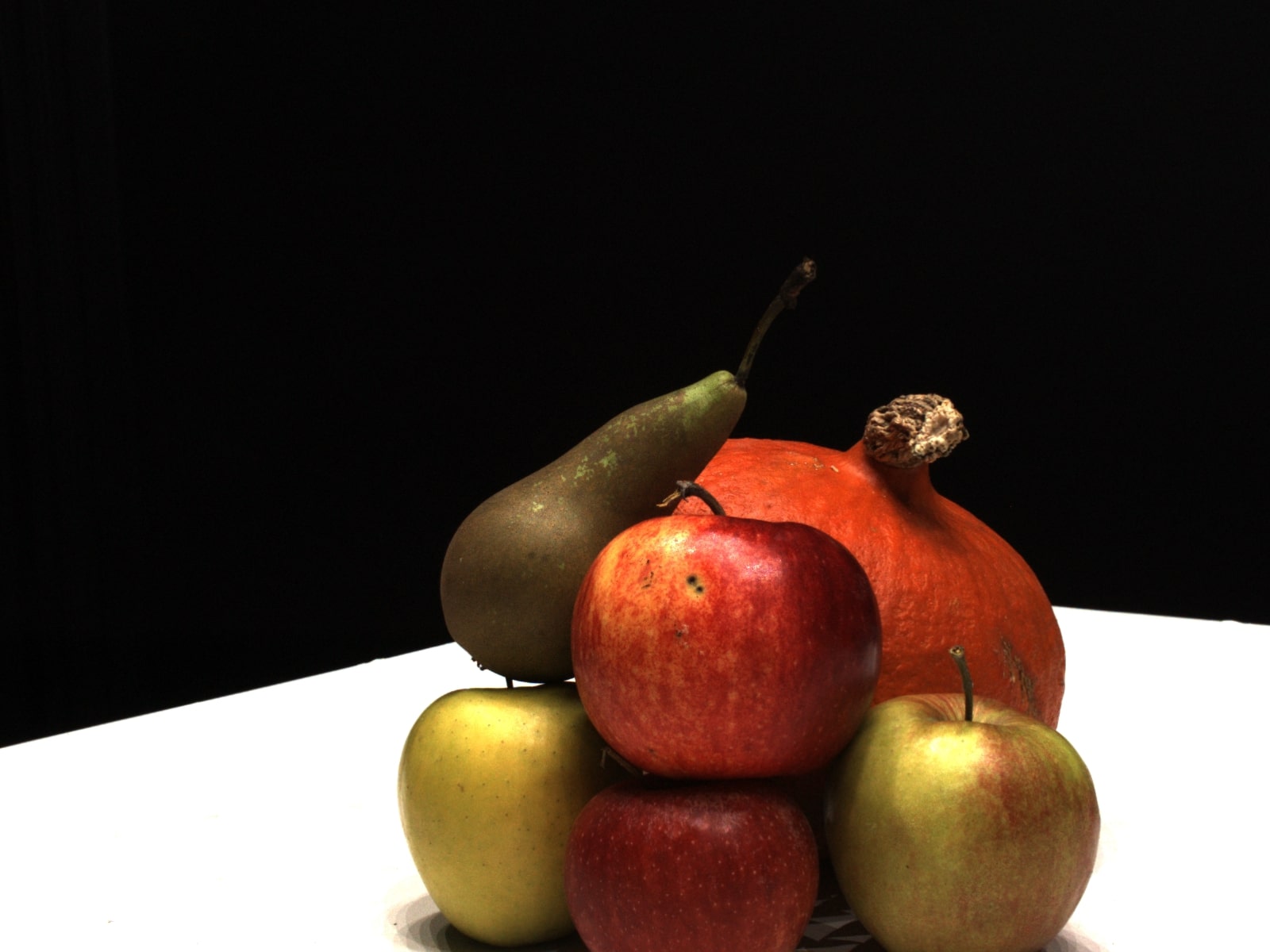}
     \end{subfigure}
     \begin{subfigure}[b]{0.19\linewidth}
         \centering
         \includegraphics[width=\textwidth]{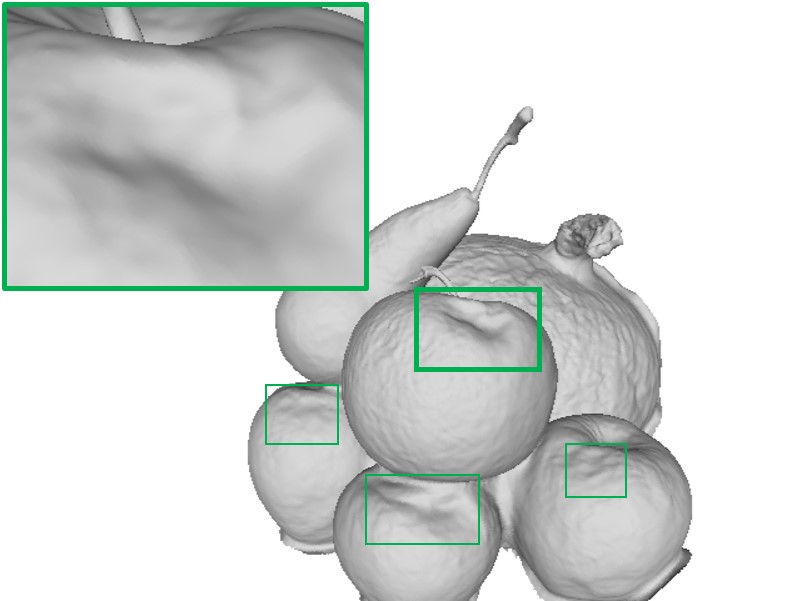}
     \end{subfigure}
     \begin{subfigure}[b]{0.19\linewidth}
         \centering
         \includegraphics[width=\textwidth]{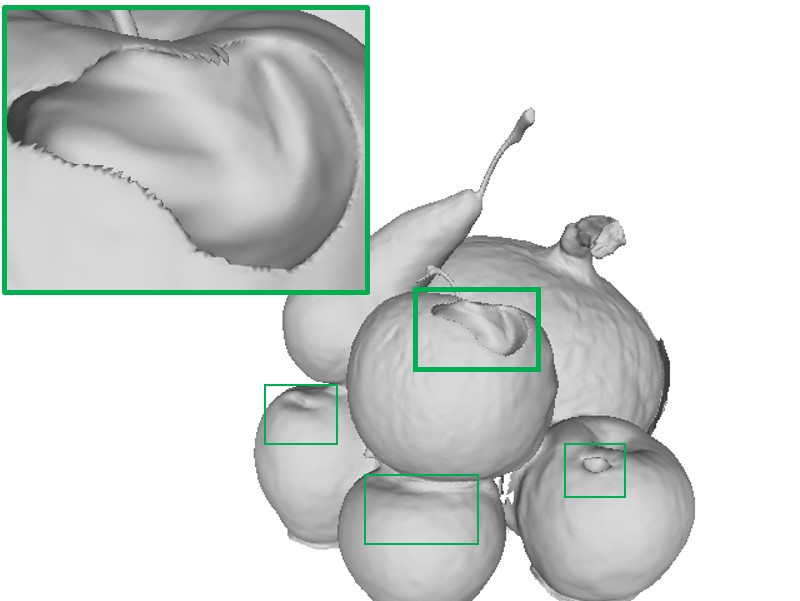}
     \end{subfigure}
     \begin{subfigure}[b]{0.19\linewidth}
         \centering
         \includegraphics[width=\textwidth]{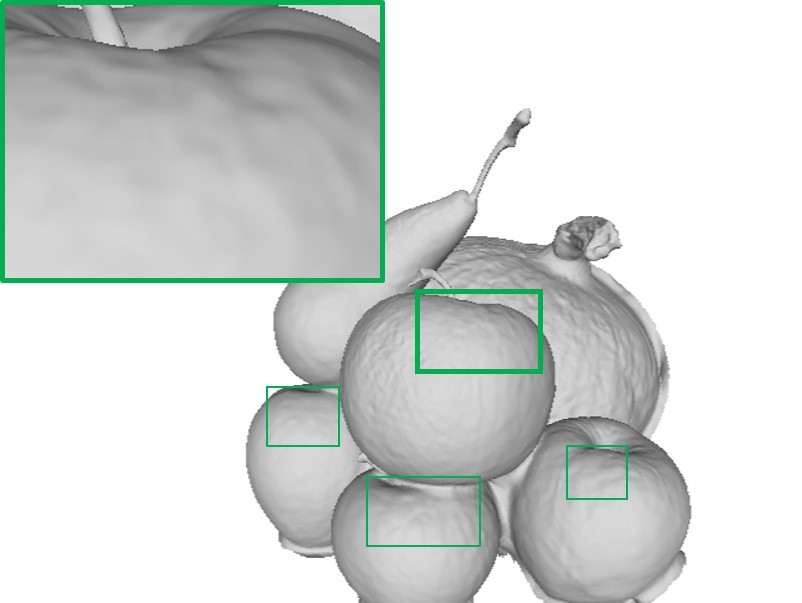}
     \end{subfigure}
     \begin{subfigure}[b]{0.19\linewidth}
         \centering
         \includegraphics[width=\textwidth]{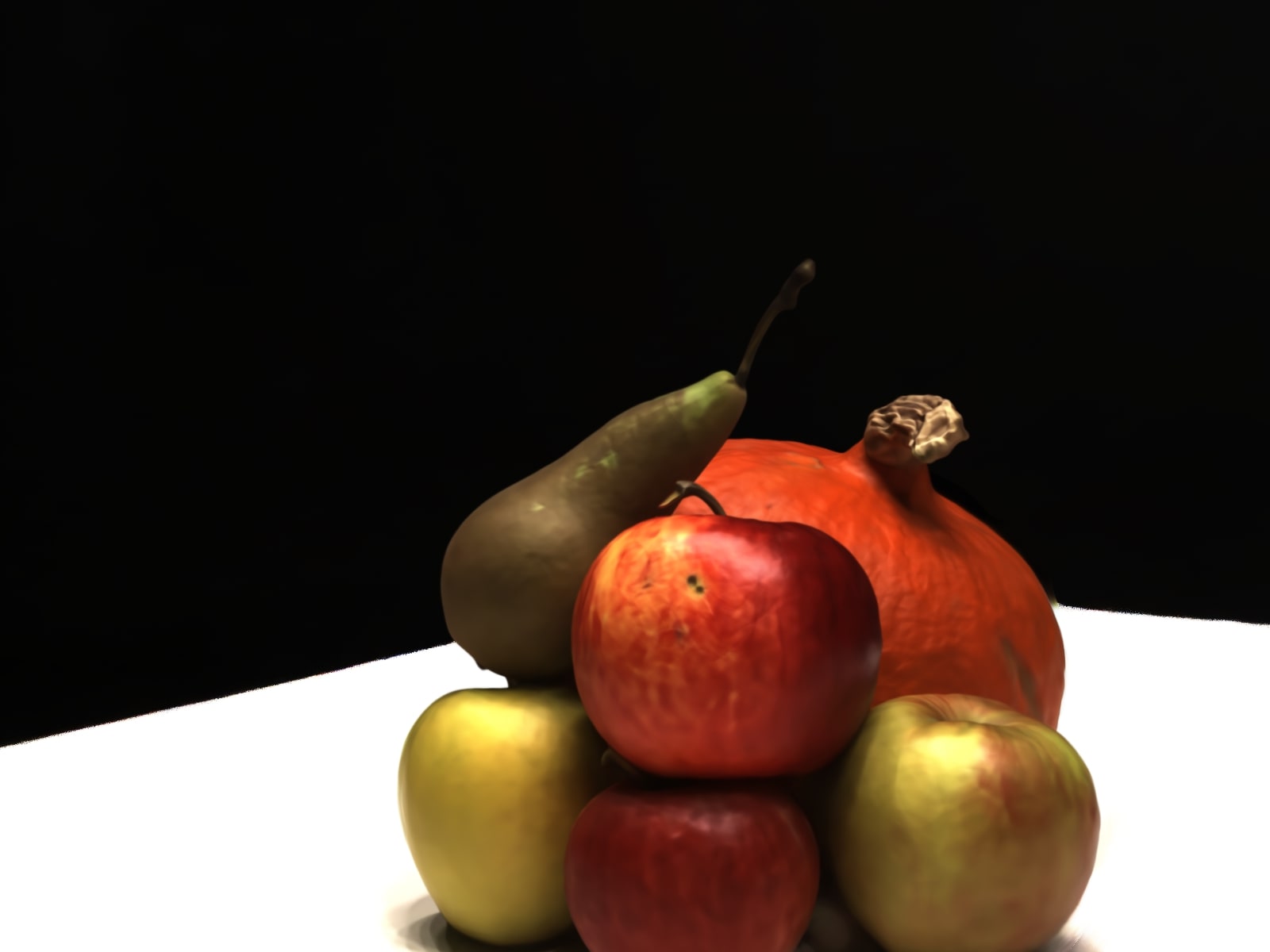}
     \end{subfigure}
    \hspace*{\fill}
    
    \vspace{1.5mm}
    
     \hspace*{\fill}
     \rotatebox{90}{\scriptsize{~~~~~~~~~~~~~~~~~~~~~~Scan 110}}
    \hspace*{\fill}
     \begin{subfigure}[b]{0.19\linewidth}
         \centering
         \includegraphics[width=\textwidth]{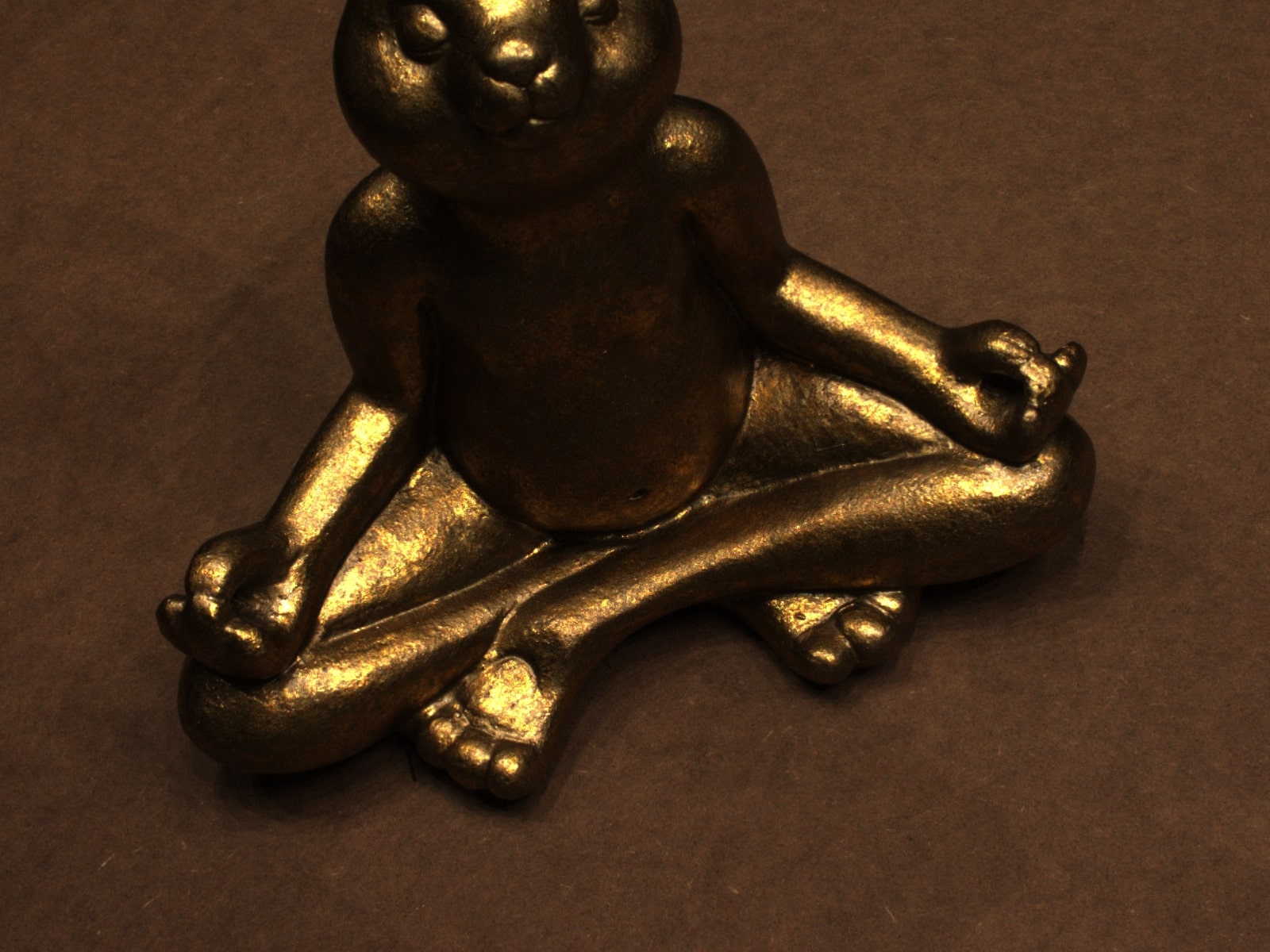}
        \caption{Reference Image}
     \end{subfigure}
     \begin{subfigure}[b]{0.19\linewidth}
         \centering
         \includegraphics[trim={0 2cm 2cm 0},clip, width=\textwidth]{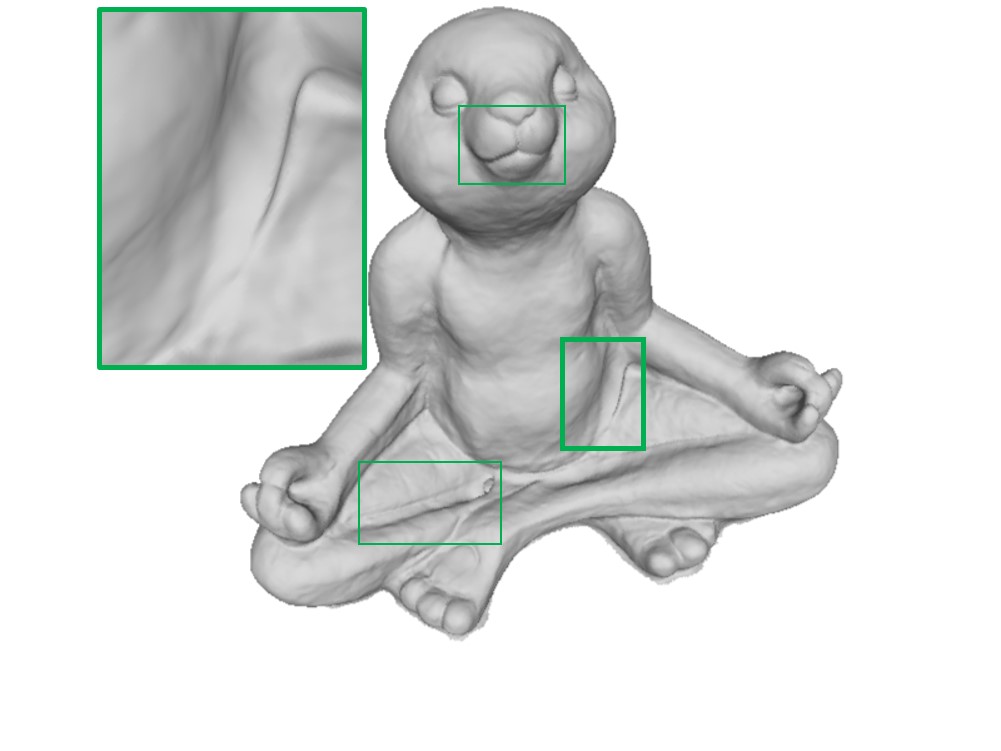}
        \caption{NeuS~\cite{neus}}
     \end{subfigure}
     \begin{subfigure}[b]{0.19\linewidth}
         \centering
         \includegraphics[trim={0 2cm 2cm 0},clip, width=\textwidth]{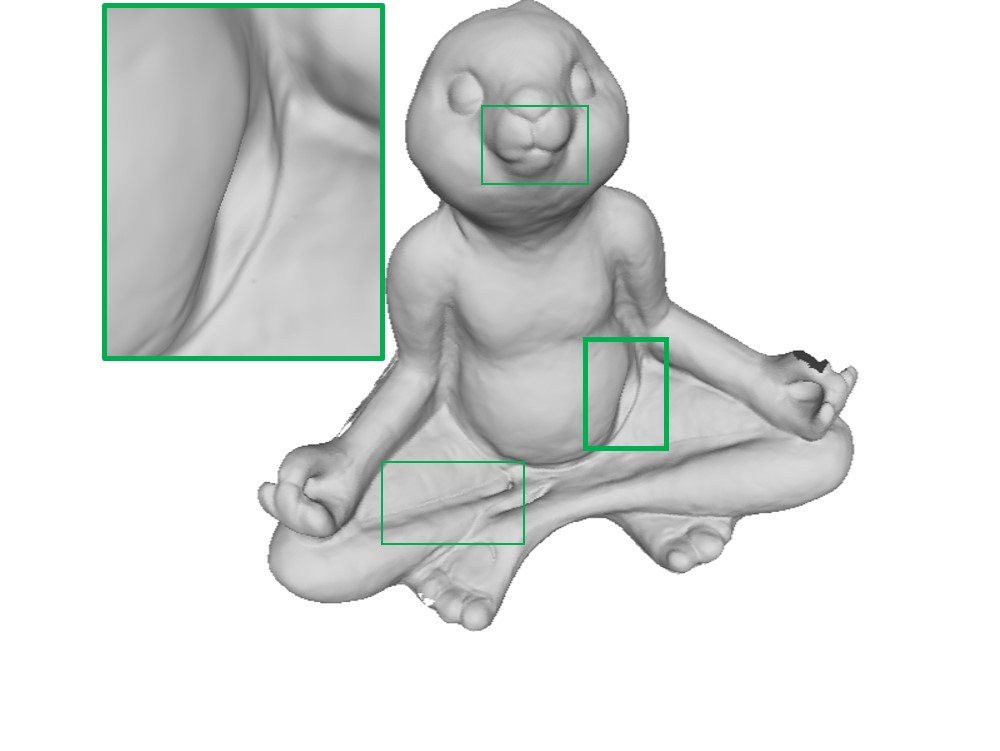}
        \caption{NeuralWarp~\cite{NeuralWarp}}
     \end{subfigure}
     \begin{subfigure}[b]{0.19\linewidth}
         \centering
         \includegraphics[trim={0 2cm 2cm 0},clip, width=\textwidth]{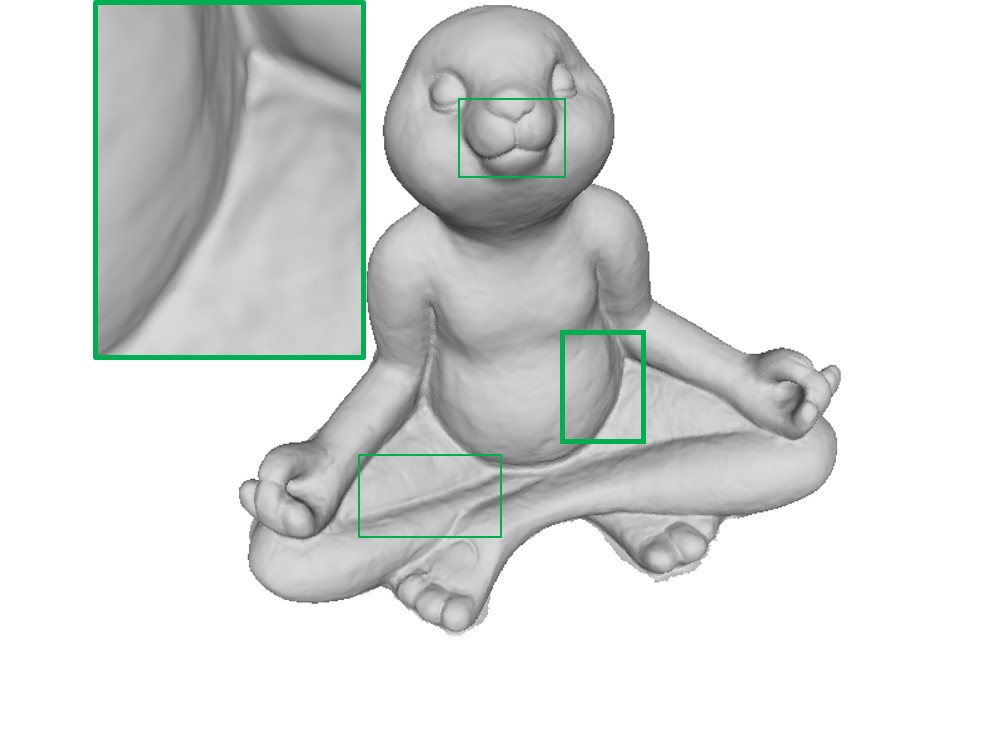}
        \caption{D-NeuS (Ours)}
     \end{subfigure}
     \begin{subfigure}[b]{0.19\linewidth}
         \centering
         \includegraphics[width=\textwidth]{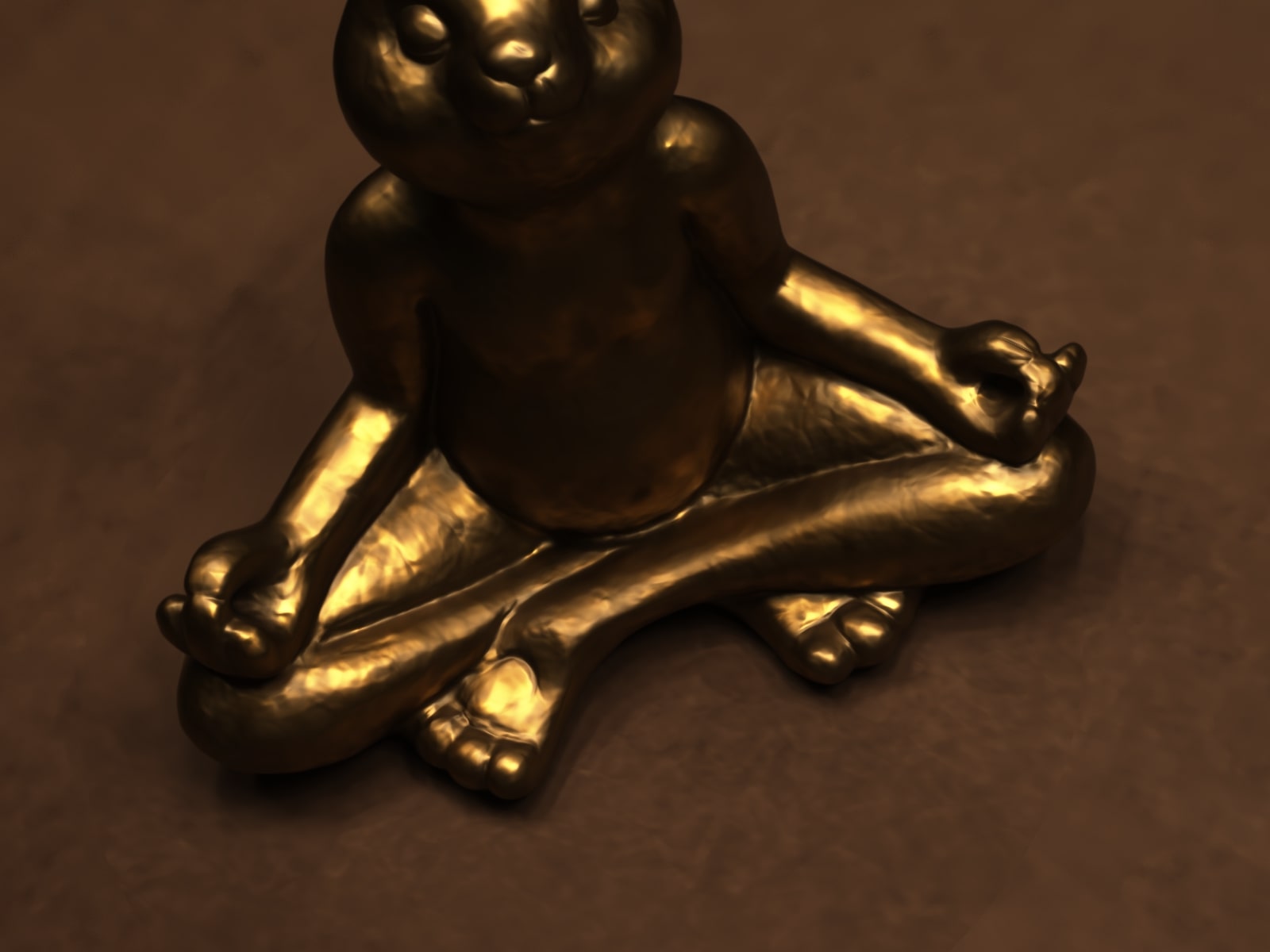}
        \caption{Our Rendered Image}
     \end{subfigure}
    \hspace*{\fill}
    \caption{Comparisons on surface reconstruction in DTU dataset.}
    \label{fig:comparison_dtu}
\end{figure*}

\begin{table*}
\begin{center}
\resizebox{\textwidth}{!}{
\begin{tabular}{c|ccccccccccccccc|c}
 Scan &24&37&40&55&63&65&69&83&97&105&106&110&114&118&122&means\\
\hline 
NeRF~\cite{nerf} & 26.24 & 25.74 & 26.79 & 27.57 & 31.96 & 31.50 & 29.58 & 32.78 & 28.35 & 32.08 & 33.49 & 31.54 & 31.00 & 35.59 & 35.51&30.65 \\
VolSDF~\cite{volsdf} & 26.28 & 25.61 & 26.55 & 26.76 & 31.57 & 31.50 & 29.38 & 33.23 & 28.03 & 32.13 & 33.16 & 31.49 & 30.33 & 34.90 & 34.75 & 30.38 \\
NeuS~\cite{neus} & \underline{28.20} & \underline{27.10} & \underline{28.13} & \underline{28.80} & \underline{32.05} & \underline{33.75} & \textbf{30.96} & \textbf{34.47} & \underline{29.57} & \underline{32.98} & \textbf{35.07} & \textbf{32.74} & \textbf{31.69} & \textbf{36.97} & \underline{37.07} & \underline{31.97}\\
D-NeuS (Ours) & \textbf{28.98} & \textbf{27.58} & \textbf{28.40} & \textbf{28.87} & \textbf{33.71} & \textbf{33.94} & \underline{30.94} & \underline{34.08} & \textbf{30.75} & \textbf{33.73} & \underline{34.84} & \underline{32.41} & \underline{31.42} & \underline{36.76} & \textbf{37.17} & \textbf{32.22}\\
\end{tabular}}
\end{center}
\caption{Qualitative results on DTU dataset in terms of PSNR, evaluating the rendering quality.}
\label{psnr}
\end{table*}

\begin{figure}[t]
     \centering
    \captionsetup[subfigure]{labelformat=empty} 
     \rotatebox{90}{\scriptsize{~~~~~~~~~~~Stone}}
    \begin{subfigure}[b]{0.315\columnwidth}
         \centering
         \includegraphics[width=\textwidth]{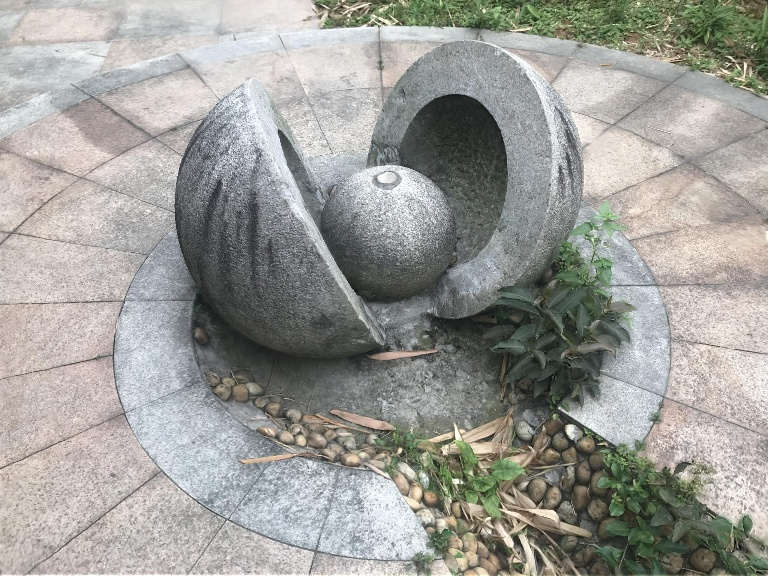}
     \end{subfigure}
     \begin{subfigure}[b]{0.315\columnwidth}
         \centering
         \includegraphics[width=\textwidth]{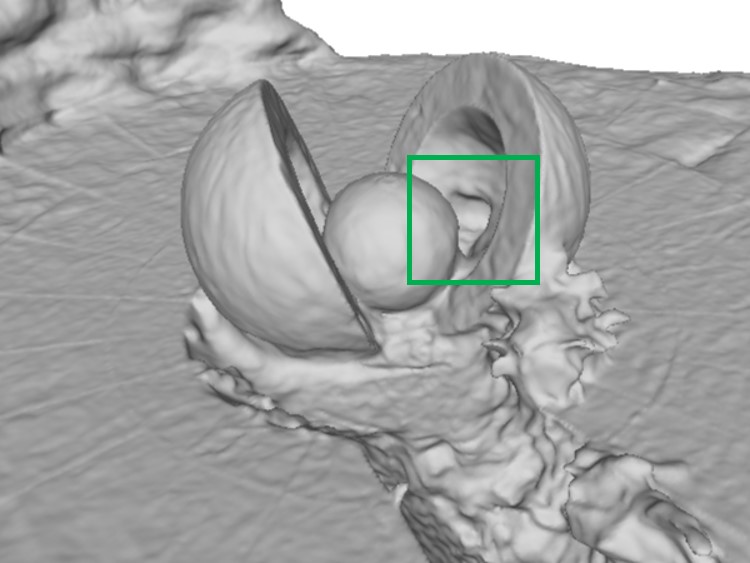}
     \end{subfigure}
     \begin{subfigure}[b]{0.315\columnwidth}
         \centering
         \includegraphics[width=\textwidth]{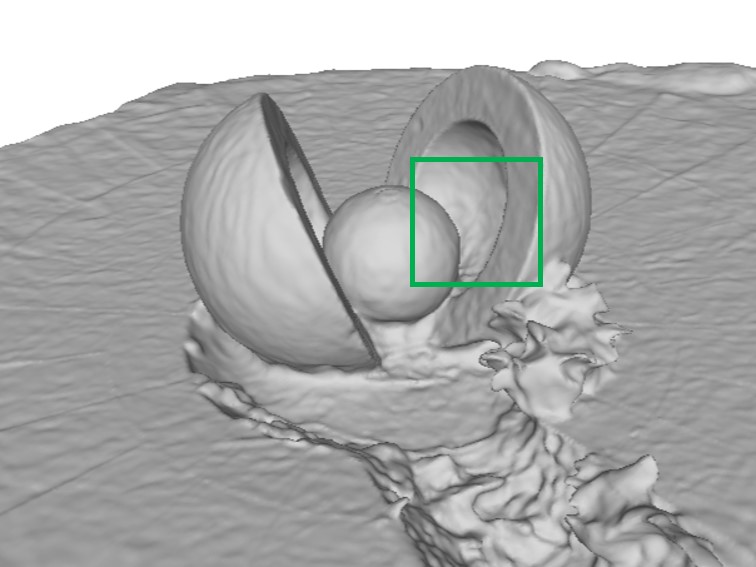}
     \end{subfigure}
     
    \vspace{1.5mm}
    
     \rotatebox{90}{\scriptsize{~~~~~~~~~~~~~~~~~~~~~Jade}}
     \begin{subfigure}[b]{0.315\columnwidth}
         \centering
         \includegraphics[width=\textwidth]{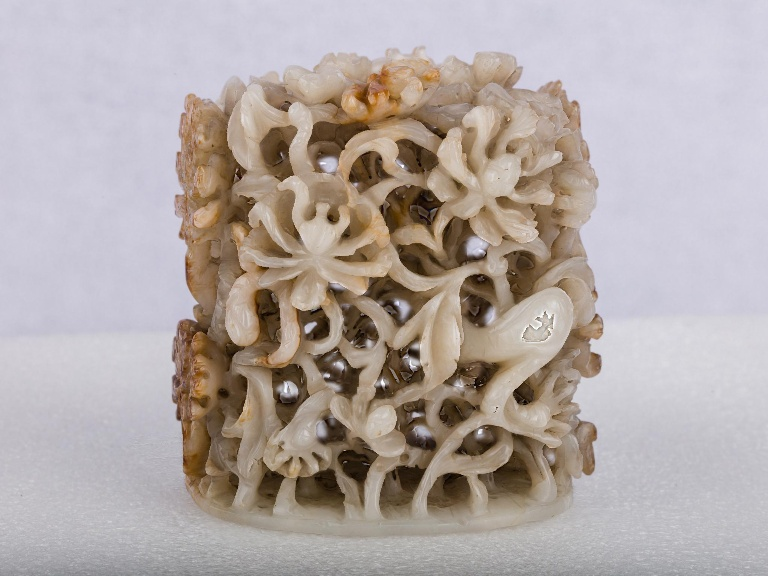}
         \caption{Reference Image}
     \end{subfigure}
     \begin{subfigure}[b]{0.315\columnwidth}
         \centering
         \includegraphics[width=\textwidth]{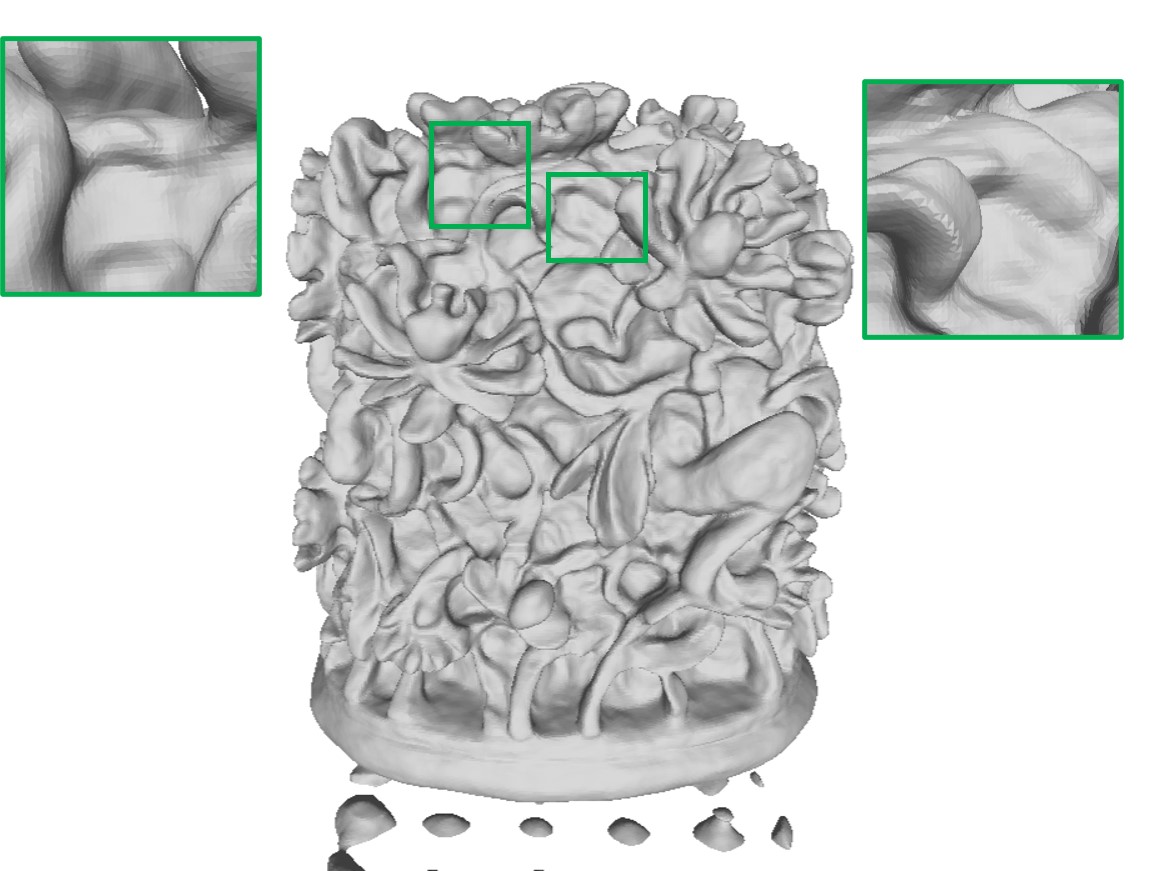}
         \caption{NeuS~\cite{neus}}
     \end{subfigure}
     \begin{subfigure}[b]{0.315\columnwidth}
         \centering
         \includegraphics[width=\textwidth]{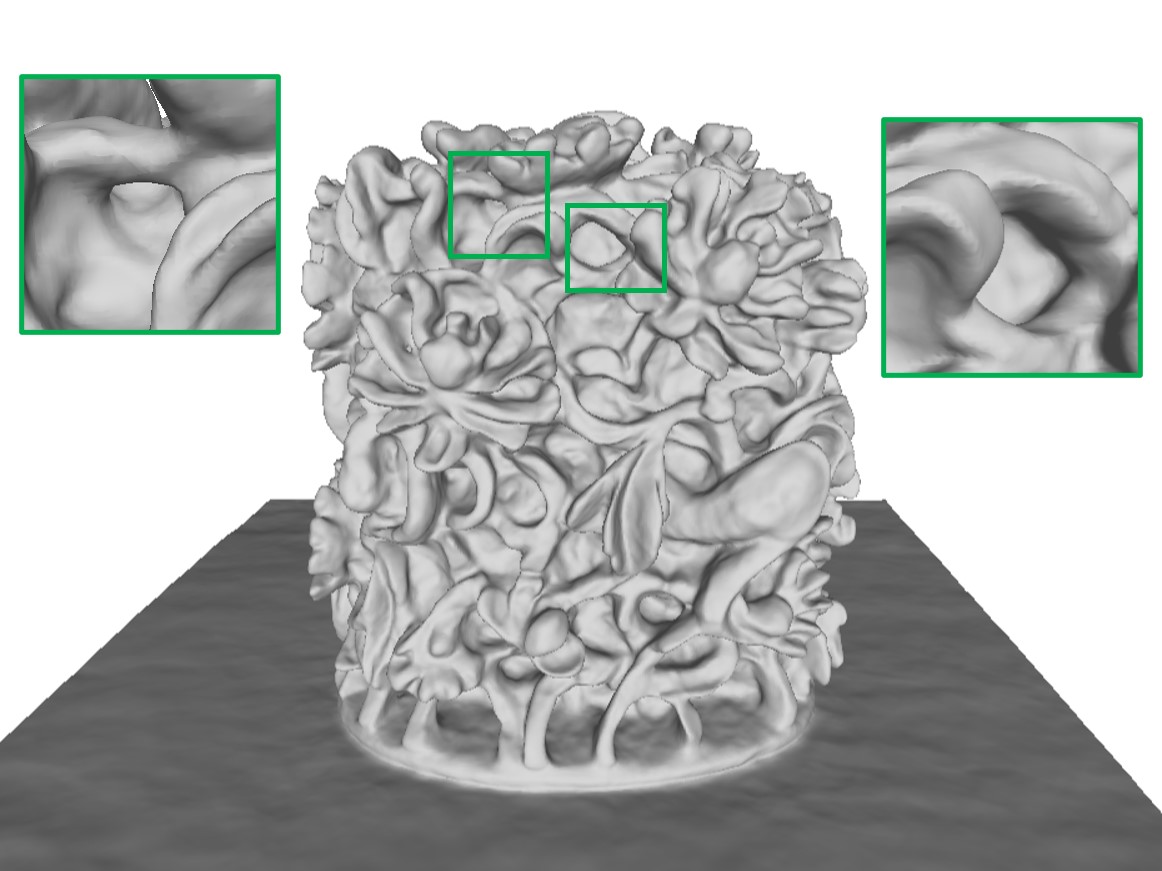}
         \caption{D-NeuS (Ours)}
     \end{subfigure}
\caption{Comparisons on surface reconstruction in BlendedMVS dataset.}
\label{fig:comparison_bld}
\end{figure}

\section{Experiments} \label{Experiments}

\subsection{Experimental Settings} \label{settings}
\noindent
\textbf{Datasets.}
To evaluate our method with full losses described in Section~\ref{loss} on the DTU dataset~\cite{dtu}, we follow previous works~\cite{idr,volsdf,neus,NeuralWarp} and select the same 15 models for comparison. Each scene contains 49 or 64 images at 1200 $\times$ 1600 resolution with camera parameters. Objects in DTU dataset have various geometries, appearances and materials, including non-Lambertian surfaces and thin structures. Furthermore, we test on 6 challenging scenes from the BlendedMVS dataset~\cite{blendedmvs}. BlendedMVS dataset provides images at a resolution of 576 $\times$ 768 with more complex backgrounds and various numbers of views vary from 24 to 143. For DTU dataset, the reconstructed surfaces are evaluated quantitatively by metrics of the Chamfer distance in mm, while we demonstrate visual comparison of the reconstruction results on BlendedMVS dataset.

\noindent
\textbf{Baselines.} 
We compare the proposed method to a traditional MVS pipeline COLMAP~\cite{colmap}, and state-of-the-art learning-based approaches: IDR~\cite{idr}, MVSDF~\cite{mvsdf}, VolSDF~\cite{volsdf}, NeuS~\cite{neus}, NeuralWarp~\cite{NeuralWarp}.

\noindent
\textbf{Network architecture.} 
Similar to \cite{idr,neus}, our geometry network $f$ is modeled by an MLP including 8 hidden layers with size of 256 and a skip connection from the input to the output of the 4-th layer. The weights of the geometry network are initialized to approximate the SDF field of a unit sphere~\cite{sdfinit}. The radiance network $c$ is an MLP consisting of 4 hidden layers MLP with 256 hidden cells. Position encoding is applied to $\mathbf{x}$ with 6 frequencies and $\mathbf{v}$ with 4 frequencies. For volume rendering, we adopt the hierarchical sampling strategy in NeuS~\cite{neus}, sampling 512 rays for each iteration, with 64 coarse and 64 fine sampled points per ray, and additional 32 points outside the unit sphere following NeRF++ \cite{zhang2020nerf++}. For multi-view features consistency, we compare each reference view with $N_{v}=2$ neighboring source views using $N_{c}=32$ feature channels.

\noindent
\textbf{Training details.}
To train our networks, we adopt Adam optimizer~\cite{kingma2014adam} using the learning rate $\text{5}e^{\text{-4}}$ with warm-up period of 5k iterations before decaying by cosine to the minimal learning rate of $\text{2.5}e^{\text{-5}}$. We initialize the trainable standard deviation for the logistic density distribution for the volume rendering weight with 0.3. We train our model for 300k iterations for 19 hours on a single NVIDIA Titan RTX graphics card. In terms of inference, rendering an image of resolution 1200 $\times$ 1600 using standard volume rendering takes approximately 7 minutes.
As for the weighting factors of losses in Eqn.~\ref{totallosss}, we fix the Eikonal weight $\alpha$ as 0.1 for the whole training. In addition, inspired by MVSDF~\cite{mvsdf}, we divide the training in three stages. In the first 50k iterations, we set the geometry bias loss weight $\beta$ as 0.01. From 50k to 150k iterations, we set $\beta$ as 0.1 and the feature consistency loss weight $\gamma$ as 0.5, while in the remaining iterations, $\beta$ and $\gamma$ are 0.01 and 0.05, respectively. After optimization, we apply Marching Cubes~\cite{lorensen1987marching} to extract a mesh from the SDF field $f$ in a predefined bounding box with the volume size of $512^3$ voxels, which takes about 57 seconds.

\subsection{Comparisons} \label{sec:Comparisons}
To evaluate the surface geometry reconstruction quality, we follow previous works and use the official evaluation code to calculate the Chamfer $L_1$ distance, which is the average of accuracy (mean distance from sampled point clouds of the reconstructed surface to the ground truth point cloud) and completeness (mean distance from the ground truth point cloud to the reconstructed counterpart). Similar to previous works~\cite{idr,unisurf,neus,volsdf, NeuralWarp}, we clean the extracted meshes with the object masks dilated by 50 pixels. Table~\ref{chamfer} shows the mean Chamfer distances of our work and baselines. Results of the baselines are reported in their original papers, except for COLMAP whose result is taken from ~\cite{neus}. Following previous works, we focus on the comparison of the approaches requiring no additional per-scene prior knowledge including object masks, depth maps or point clouds. As illustrated in Table~\ref{chamfer}, our method surpasses the baselines by a noticeable margin and achieve the lowest mean Chamfer distance.

Fig.~\ref{fig:comparison_dtu} qualitatively compares the reconstructed surface geometry of our method and the baselines on DTU~\cite{dtu} dataset. Surfaces from NeuS~\cite{neus} are more noisy and bumpy, especially in the plane (Scan 40) or smooth region (Scan 63 and 110), while NeuralWarp struggles to reconstruct the surface boundaries (Scan 40 and 110) and non-Lambertian area (Scan 63). In contrast, our method is robust to these challenging cases, recovering fine geometry details with high accuracy and fidelity.
In addition to surface geometry, D-NeuS also achieves photorealistic image rendering. As reported in Table~\ref{psnr}, we quantitatively evaluate the PSNR of rendering results from our method, which outperforms other state-of-the-art methods. Following previous works, we only evaluate the PSNR of the pixels inside the object masks provided by IDR~\cite{idr}.


Qualitative results on BlendedMVS dataset~\cite{blendedmvs} are shown in Figure~\ref{fig:comparison_bld}. Our method is robust to challenging surfaces, such as the seriously occluded area in Stone, as well as highly complex concave holes in Jade, while NeuS struggles to recover the fine geometric details of these complicated surfaces.

\begin{figure}[t]
     \centering
    \captionsetup[subfigure]{labelformat=empty} \begin{subfigure}[b]{0.47\columnwidth}
         \centering
         \includegraphics[width=\textwidth]{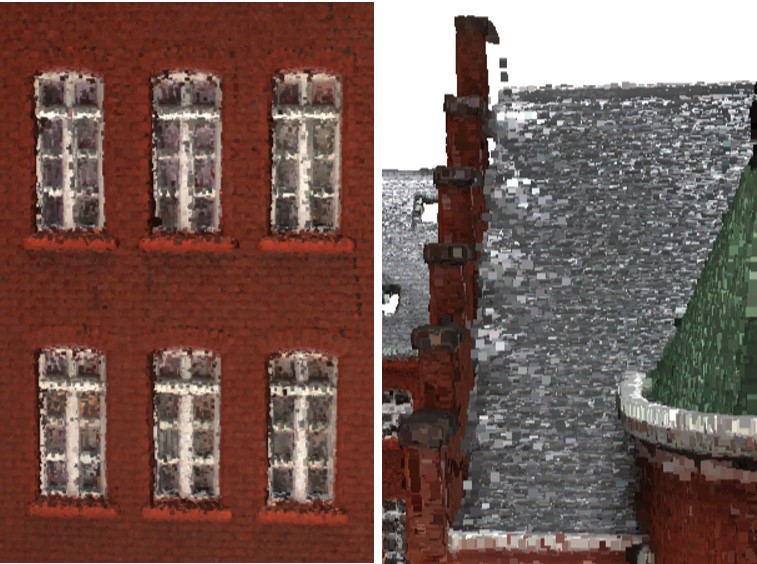}
         \caption{Reference Point Cloud}
         \label{fig:pc}
     \end{subfigure}
    
    \vspace{1mm}
    
    \hspace*{\fill}
     \begin{subfigure}[b]{0.47\columnwidth}
         \centering
         \includegraphics[width=\textwidth]{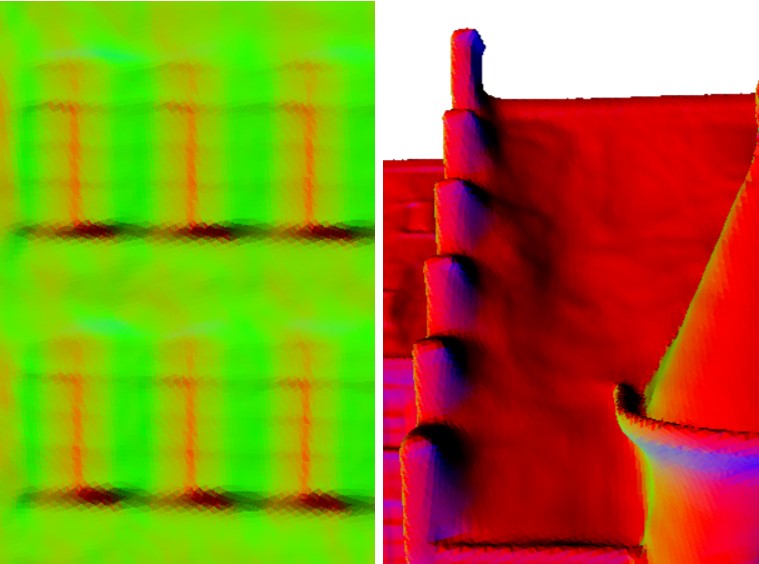}
         \caption{Baseline}
         \label{fig:Baseline}
     \end{subfigure}
    \hspace*{\fill}
     \begin{subfigure}[b]{0.47\columnwidth}
         \centering
         \includegraphics[width=\textwidth]{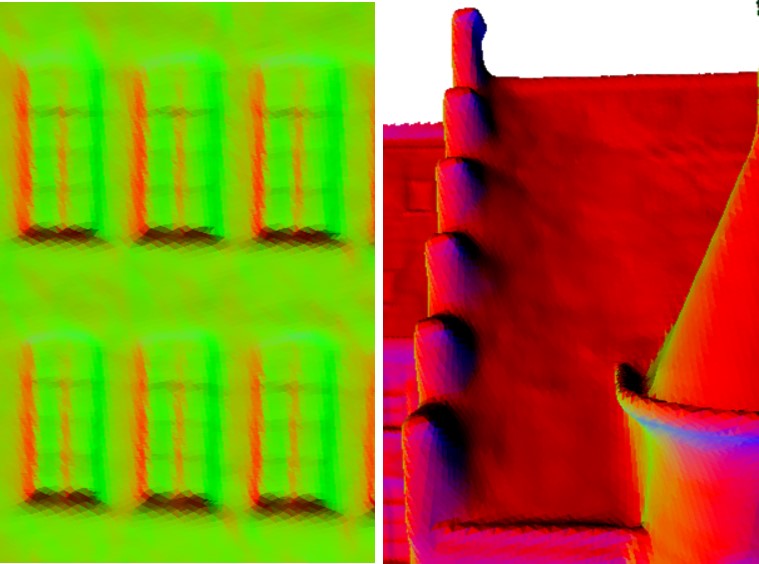}
         \caption{W/ bias}
         \label{fig:wbias}
     \end{subfigure}
    \hspace*{\fill}
    
    \vspace{1mm}
    
    \hspace*{\fill}
     \begin{subfigure}[b]{0.47\columnwidth}
         \centering
         \includegraphics[width=\textwidth]{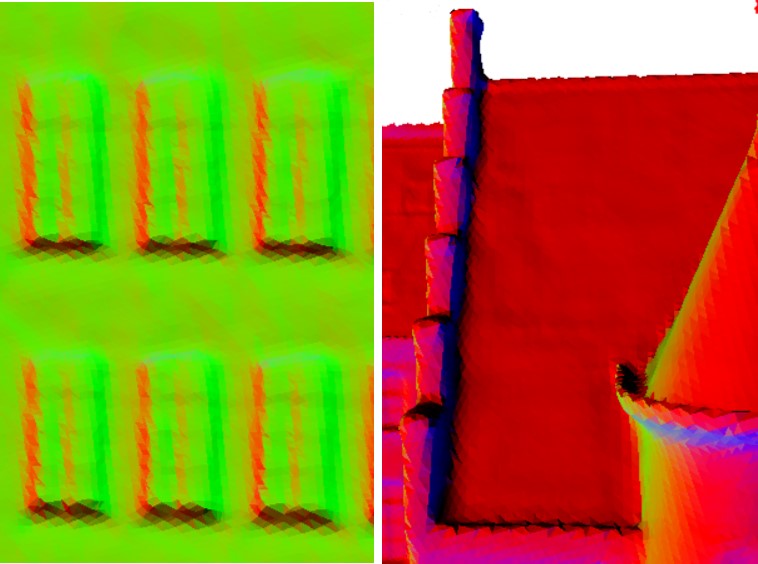}
         \caption{W/ feature}
         \label{fig:wfeat}
     \end{subfigure}
    \hspace*{\fill}
     \begin{subfigure}[b]{0.47\columnwidth}
         \centering
         \includegraphics[width=\textwidth]{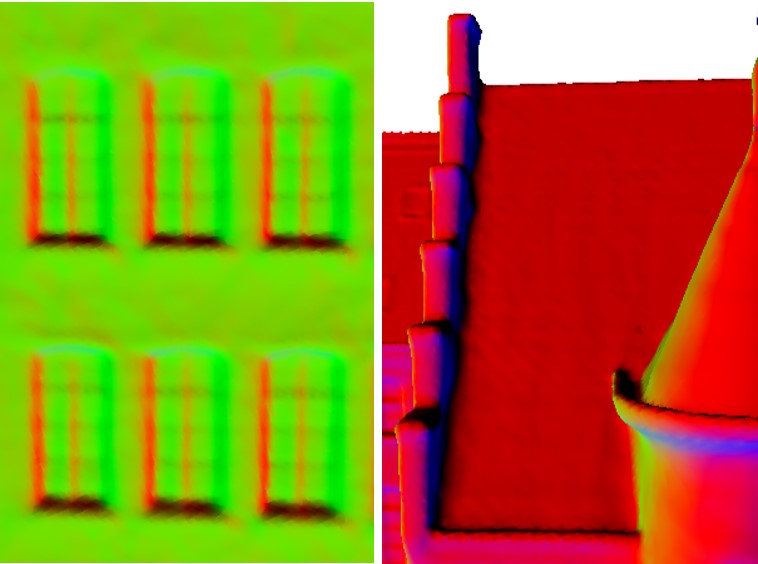}
         \caption{Ours}
         \label{fig:ours}
     \end{subfigure}
    \hspace*{\fill}
\caption{Qualitative results of ablation study on DTU dataset. To better illustrate the geometry details, we visualize the surface normals of two challenging regions.}
\label{ablation_fig}
\end{figure}

\begin{table}[t]
\begin{center}
\begin{tabular}{c|ccc|c}
  & $\mathcal{L}_{color}$ &  $\mathcal{L}_{bias}$ & $\mathcal{L}_{feature} $& Mean Chamfer\\
\hline 
Baseline & \checkmark& & & 0.84 \\ 
W/ bias & \checkmark & \checkmark & &0.76 \\ 
W/ feature & \checkmark & &\checkmark&0.63 \\
Ours & \checkmark & \checkmark & \checkmark& 0.61\\
\end{tabular}
\end{center}
\caption{Quantitative results of ablation study on DTU dataset.}
\label{ablation_table}
\end{table} 

\subsection{Ablation Study}  \label{sec:ablation}
We evaluate different components of our method by an ablation study on the DTU dataset. Specifically, we use NeuS~\cite{neus} as the baseline on which we build D-NeuS, and progressively combine our proposed losses. As demonstrated in Fig.~\ref{ablation_fig}, the geometry bias loss successfully recovers the fine geometric details of the windows, while multi-view feature consistency loss faithfully reconstructs the surface boundary: the connecting part between the roof and the facade. 

Table~\ref{ablation_table} shows the quantitative result using the mean Chamfer distance. Both contributions of our work can improve the surface reconstruction, and D-NeuS combine their advantages for the best performance.

\section{Discussion}
\textbf{Limitations.}
Similar to other neural implicit surface reconstruction methods, training our model takes some hours for each scene. In addition, the rendered images struggle to recover the high-frequency patterns in the input images. Lastly, a certain number of dense input views are required for high-quality reconstruction.

\textbf{Future works.}
One interesting future direction is to represent the scene with a multi-resolution structure, \eg, Instant-ngp~\cite{muller2022instant}, for fast optimization as well as high-frequency local details. Moreover, it is promising to generalize the reconstruction to new scenes using learned image priors like~\cite{yu2021pixelnerf} or geometry priors like~\cite{johari2022geonerf}, which may also enable surface reconstruction from a sparse set of views.

\textbf{Conclusions.}
We introduce D-NeuS, a volume rendering-based neural implicit surface reconstruction method recovering fine-level geometric details. We analyze the cause for geometry bias between the SDF field and the volume rendered color, and propose a novel loss function to constrain the bias. In addition, we apply multi-view feature consistency to surface points derived by interpolating the zero-crossing from sampled SDF values. Extensive experiments on different datasets show that D-NeuS is able to reconstruct high-quality surfaces with fine details and outperforms the state of the art both qualitatively and quantitatively. 

\section*{Acknowledgement}
This work has partly been funded by the H2020 European project Invictus under grant agreement no. 952147 as well as by the Investitionsbank Berlin with financial support by European Regional Development Fund (EFRE) and the government of Berlin in the ProFIT research project KIVI.

{\small
\bibliographystyle{ieee_fullname}
\bibliography{egbib}
}

\end{document}